\documentclass{article} 
\usepackage{iclr2027_conference,times}

\usepackage{amsmath,amsfonts,bm}

\def\eqref#1{equation~\ref{#1}}

\def\1{\bm{1}}

\DeclareMathAlphabet{\mathsfit}{\encodingdefault}{\sfdefault}{m}{sl}
\SetMathAlphabet{\mathsfit}{bold}{\encodingdefault}{\sfdefault}{bx}{n}

\usepackage{hyperref}
\usepackage{url}
\usepackage{booktabs}
\usepackage{multirow}
\usepackage{graphicx}
\usepackage{xcolor}
\usepackage[T1]{fontenc}
\usepackage{textcomp}
\usepackage{listings}
\usepackage[most]{tcolorbox}
\newtcolorbox{takeaway}{enhanced, colback=gray!4, colframe=gray!45, boxrule=0.8pt,
  arc=2mm, left=7pt, right=7pt, top=5pt, bottom=5pt, before skip=8pt, after skip=10pt}
\title{What Should the Reflector See? \\ An Empirical Study of Evidence in Reflective Prompt Optimization}

\author{Xiaofan Zhou \qquad Lu Cheng \\
Pennsylvania State University \\
\texttt{xpz5429@psu.edu} \qquad \texttt{lqc5822@psu.edu}}
\hypersetup{
  hidelinks,
  pdftitle={What Should the Reflector See? An Empirical Study of Evidence in Reflective Prompt Optimization},
  pdfauthor={Xiaofan Zhou and Lu Cheng}
}

\iclrfinalcopy
\begin{document}

\maketitle
\pagestyle{plain}
\thispagestyle{plain}

\begin{abstract}
Reflective prompt optimization revises instructions using examples of a
model's behavior, but which evidence the reflector should receive remains
unclear. We study evidence composition, visibility of examples, candidate
selection and domain-knowledge policy within a single-parent Pareto-guided
search. Using Qwen3.5-9B as both task model and reflector, we evaluate nine
reflection strategies on five datasets. From these experiments, we find
three distinct patterns. For performance improvement, Failures-only produces
the largest mean test gain (+8.0 percentage points), while Balanced-mix and
No-examples+Val share the best mean performance rank. For reflection
effectiveness, No-examples achieves the best rank for improving sampled
parents, yet yields only a 1.4-point mean test gain: local reflection success
does not necessarily produce a stronger final prompt. For overfitting
assessment, Failures-only and Balanced-mix share the lowest mean
calibration-gap rank, while larger gaps on GPQA and IFBench show that
calibration gains can overstate held-out improvement. This gap is a
descriptive indicator, not a direct measure of overfitting. Together, these
results show why reflection strategies should be assessed separately on
final performance, parent improvement and calibration-to-test transfer.
\end{abstract}

\begin{figure}[!ht]
\centering
\includegraphics[width=\textwidth]{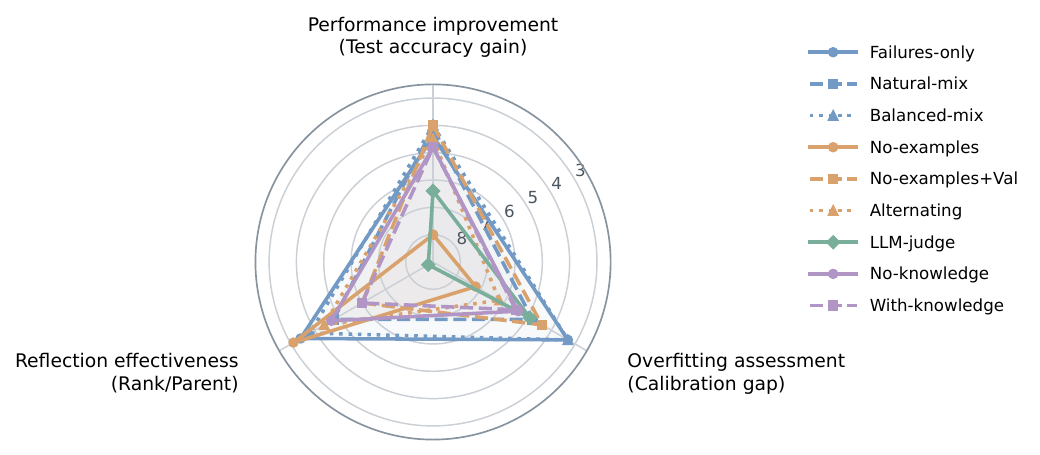}
\caption{Three-dimensional strategy profiles: mean within-dataset ranks over
five datasets, with ties sharing ranks. All axes use the same linear, zoomed
range 2.5--9; outer is better. Test gain ranks high-to-low; calibration gap
(calibration gain minus test gain) ranks low-to-high. Reflection effectiveness uses
Rank/Parent from Table~\ref{tab:iterations}: lower rounds per parent improvement
rank better, with zero-improvement runs assigned 20 rounds. Calibration gap
is descriptive, not a direct measure of overfitting. Colors encode evidence
composition (blue), example visibility (orange), judging (green) and knowledge
policy (purple); line styles distinguish strategies.}
\label{fig:overview}
\end{figure}

\section{Introduction}
\label{sec:intro}

Large language models (LLMs) have become capable general-purpose problem
solvers \citep{brown2020language}, but how much of this capability users
actually obtain often depends on the prompt, especially on hard reasoning
tasks \citep{kojima2022large}.
Writing a prompt that works well often takes an experienced prompt engineer
and many rounds of trial and error \citep{zhou2023ape}, which is out of reach
for many non-expert users.
A common way to improve performance is to fine-tune the model, for example
with reinforcement learning such as GRPO \citep{shao2024deepseekmath}, but
this requires access to the weights and substantial compute.
To address this, a series of training-free methods, including prompt
evolution, have been proposed: they keep the model fixed and let an LLM
automatically propose, test and refine the prompt
\citep{zhou2023ape,yang2024opro,pryzant2023protegi,agrawal2025gepa}.

Most of these methods follow the same iterative procedure.
They maintain a set of candidate prompts and, in each round, run a selected
prompt on a small batch of training examples.
An LLM then reads the results, including the inputs, the model's responses and
feedback on them, and writes a revised prompt.
The revised prompt is evaluated on held-out examples and kept if it improves
performance, so the search gradually accumulates better prompts.
The key step in this loop is the revision: every improvement is inferred from
the few examples the LLM reads in that round, which makes the choice of these
examples central to what the method can learn.

Prompt evolution can improve performance without updating any weights and
with a modest number of rollouts.
Recent methods advance the search procedure, including parent selection and
candidate maintenance and combination \citep{guo2024evoprompt,agrawal2025gepa}.
Evidence design has also received targeted analysis: StraGo
\citep{wu2024strago} separately ablates positive and negative experiences,
while OPRO \citep{yang2024opro} compares different numbers of exemplars,
including none, and studies the effect of showing instruction scores.
These analyses motivate a broader comparison of evidence composition,
example visibility, candidate selection and knowledge policy within one
search backbone. We compare these choices across five datasets,
including whether the reflector should retain task-specific knowledge or
write generic strategies. Our no-examples control sees only the current
instruction and task description; it differs from OPRO's no-exemplar
condition, which retains historical instructions and their scores.

To address these limitations, we compare nine reflection strategies using
Qwen3.5-9B as both task model and reflector. We vary evidence composition,
whether generation sees examples, how candidates are selected, and whether
reflection explicitly retains domain knowledge. The strategies share a
single-parent Pareto-guided search and iteration budget, while their realized
evaluation costs can differ.
We study five datasets spanning mathematics, graduate-level science,
multi-hop question answering and instruction following, reporting test
accuracy averaged over three final evaluations.
No strategy is best on all tasks. Failures-only has the largest average gain,
driven by LiveBench-Math, while Balanced-mix and No-examples+Val share the
best mean accuracy rank.
The effects of balancing evidence and retaining knowledge vary substantially
by dataset, motivating task-specific validation rather than a universal
reflection recipe.

\section{Related Work}
\label{sec:related}

\textbf{Prompt optimization} improves LLMs without changing their weights, but
good prompts, such as chain-of-thought prompting \citep{wei2022chain}, have
traditionally been found by hand.
To automate this, recent methods let an LLM propose and refine prompts: APE
\citep{zhou2023ape} induces instructions from demonstrations and keeps the
highest-scoring ones, OPRO \citep{yang2024opro} shows the LLM earlier prompts
with their scores and asks for a better one, and GEPA
\citep{agrawal2025gepa} reflects on execution traces and textual feedback.
These works each propose a new optimizer and compare it with others as a
complete system; we instead take one optimizer as given and study a single
component of it, the evidence its reflector reads.

\textbf{Evolutionary algorithms} have been used to search over prompts:
EvoPrompt \citep{guo2024evoprompt} lets the LLM act as the crossover and
mutation operators on a population of prompts, GEPA keeps a Pareto front of
candidates so that complementary improvements are not lost, and C-Evolve
\citep{li2025cevolve} evolves a group of prompts whose outputs are aggregated
by majority voting.
AlphaEvolve \citep{novikov2025alphaevolve} applies the same LLM-driven
evolutionary loop to code, using automated evaluators to discover improved
algorithms.
This line of work has mainly advanced the search procedure, such as how
parents are selected and how candidates are maintained and combined; we keep
the search procedure fixed and vary what the LLM sees when it proposes a new
candidate.

\textbf{Feedback-driven improvement} uses verbal feedback instead of gradient
updates. Reflexion \citep{shinn2023reflexion} stores reflections on task
feedback in episodic memory to improve subsequent trials, whereas Self-Refine
\citep{madaan2023selfrefine} directly revises outputs using self-generated
critiques.
For prompts, ProTeGi \citep{pryzant2023protegi} criticizes a prompt on the
examples it gets wrong and edits it along this ``textual gradient'', and
TextGrad \citep{yuksekgonul2024textgrad} propagates such feedback through
multi-component systems.
StraGo \citep{wu2024strago} observes that revising a prompt only from its
failures can break cases it previously solved, and derives strategies from
both successful and failed cases, with separate ablations of positive and
negative experiences. Building on such targeted analyses, we compare failures
only, natural and balanced mixes, no examples, and restrictions on what the
reflector may write under the same search procedure and iteration budget.

\textbf{Compound AI systems and agents} extend prompt optimization to programs
with several LLM calls.
DSPy \citep{khattab2024dspy} bootstraps few-shot demonstrations for each
module, MIPRO \citep{opsahlong2024mipro} jointly optimizes instructions and
demonstrations with Bayesian optimization, and MASPOB \citep{hong2026maspob}
optimizes the prompts of a multi-agent system with a bandit that models the
coupling between agents.
Reflective optimizers such as GEPA and TextGrad revise each module from
evidence gathered on that module, so the question we study arises for every
module they update; we study it in the single-prompt setting, where its
effect can be isolated.

\section{Method}
\label{sec:method}
\subsection{Single-parent Pareto-guided search}
\label{sec:loop}
Our backbone is a Pareto-guided prompt search inspired by GEPA
\citep{agrawal2025gepa}, not an unchanged implementation of its optimizer.
The state contains all generated prompt texts and a subset with per-example
calibration scores. A prompt is on the strict Pareto front if no other
calibrated prompt is at least as good on every calibration example and
strictly better on at least one. Each iteration samples one parent from this
front with probability proportional to its mean calibration score; all-zero
weights fall back to uniform sampling.

The parent answers 20 randomly selected training examples. The strategy
selects five groups of three from these outputs, and one reflection call per
group generates one child. Groups are selected independently and may overlap.
Empirical strategies evaluate each child on its own group, compare its score
with the parent's score on the same examples, and choose the highest-scoring
strictly improved child, breaking ties by improvement. If none improves,
calibration is skipped. Eligible children receive a full calibration evaluation;
only calibrated prompts can subsequently be parents or the final output.
Parseable child texts are retained in the pool even if not calibrated.
After 20 iterations, the highest-calibration-mean prompt is returned, with
ties favoring the earlier candidate. The held-out test set is not consulted
during this process.

\subsection{Reflection strategies}
\label{sec:strategies}
We vary evidence composition, visibility of examples, candidate selection and
knowledge policy. All strategies share the backbone and iteration budget;
they do not necessarily use the same number of model calls.
\begin{description}
\item[Failures-only] draws each group only from parent failures. If the batch
has only one or two failures, these are repeated to fill three entries; if
there are none, the iteration skips reflection and calibration.
\item[Natural-mix] samples groups uniformly from the evaluated batch.
\item[Balanced-mix] targets equal proportions of failures and successes:
with three examples it randomly targets either a 2:1 or a 1:2 split.
When one class is insufficient, the other class fills the remaining slots.
The preceding three strategies use empirical own-group selection and the
strict local improvement gate.
\item[No-examples] generates five prompts using only the task description
and current instruction. It sends the first candidate directly to calibration
if it is parseable, without three-example
validation or a local improvement gate. If that first candidate is unparseable,
calibration is skipped. The common parent batch is computed but its examples
are not used for this generation or selection.
\item[No-examples+Val] generates prompts without examples, then uses empirical
selection and the local gate on naturally sampled three-example groups.
\item[Alternating] generates without examples on odd reflection calls and
with naturally sampled evidence on even calls. Both types of call use
empirical selection and the local gate; the blind step is therefore not the
same selection procedure as No-examples.
\item[LLM-judge] uses natural evidence for generation and adapts the
LLM-as-a-judge paradigm \citep{zheng2023judging} to candidate selection.
The judge sees candidate texts and evidence and selects one child, which is
then tested on its own group for the local improvement gate. Empirical child
scores are not shown to the judge.
\item[No-knowledge / With-knowledge] use natural evidence and empirical
selection. Their reflection prompts respectively prohibit or request
domain facts, formulas and mechanisms inferred from the examples.
Both allow procedures and output conventions. Other data-grounded strategies
use a neutral template with no instruction about including or excluding
domain-specific knowledge.
\end{description}

\section{Experimental Setup}
\label{sec:setup}

\subsection{Research questions}
\label{sec:research-questions}
Our experiments address three questions:
\begin{enumerate}
\item \emph{RQ1: Which reflection strategies improve held-out performance,
and is any strategy consistently best across tasks?} We compare the nine
strategies by test accuracy, improvement over the initial prompt and
within-dataset rank.
\item \emph{RQ2: Which reflection choices drive these differences?} We
examine evidence composition, using examples for generation or selection,
empirical selection versus judging, and explicit knowledge retention.
\item \emph{RQ3: How reliably do local improvements advance the search?}
We measure how often calibrated children surpass their parent or the
previous best prompt, how many search rounds these improvements require,
and how calibration gains relate to held-out gains.
\end{enumerate}

\subsection{Datasets}
\label{sec:datasets}
We evaluate nine strategies on five datasets spanning four domains
(Table~\ref{tab:datasets}). LiveBench-Math \citep{white2025livebench} uses
the AMPS-Hard and competition subsets. OlympiadBench \citep{he2024olympiadbench}
uses the English, text-only, open-ended competition mathematics subset
(\texttt{OE\_TO\_maths\_en\_COMP}). Mathematical answers are scored by symbolic
equivalence. GPQA-main \citep{rein2024gpqa} measures graduate-level science
question answering by the selected option. IFBench \citep{pyatkin2025ifbench}
is scored by the fraction of automatically verified constraints satisfied;
its accuracy is therefore not the fraction of fully correct responses.

For HotpotQA \citep{yang2018hotpotqa}, we use the distractor setting but expose
only the three highest-ranked of each question's ten candidate paragraphs.
BM25 \citep{robertson2009bm25} uses $k_1=1.2$, $b=0.75$ and corpus-wide
inverse document frequencies.
Retrieval is computed once and cached across strategies, so prompt evolution
changes the reader instruction, not the retrieved context. Answers are scored
by normalized exact match. This retrieve-then-read construction differs from
giving the model all ten candidate paragraphs.

Training, calibration and test examples are
disjoint, with default sizes 100/150/150; smaller datasets are split into three
equal parts. The completed grid has $5\times9=45$ runs.
All strategies use the same examples and initial instruction within a
dataset.

\begin{table}[t]
\centering
\caption{Datasets. Sizes are training / calibration / test.}
\label{tab:datasets}
\begin{tabular}{llll}
\toprule
Dataset & Domain & Splits & Metric \\
\midrule
LiveBench-Math & competition math & 98 / 98 / 98 & equivalence \\
OlympiadBench & competition math & 100 / 150 / 150 & equivalence \\
GPQA & graduate-level science & 100 / 150 / 150 & option accuracy \\
HotpotQA & multi-hop QA & 100 / 150 / 150 & exact match \\
IFBench & instruction following & 100 / 100 / 100 & constraints satisfied \\
\bottomrule
\end{tabular}
\end{table}

\subsection{Implementation and search budget}
\label{sec:impl}
We use Qwen3.5-9B \citep{qwen35} for task execution, reflection and judging,
served with vLLM \citep{kwon2023vllm} and thinking mode disabled.
The search runs for 20 iterations. In each iteration, one calibrated parent
is sampled from the strict Pareto front, weighted by its mean calibration
score, and evaluated on 20 training examples. Five groups of three examples
are drawn independently from these results and yield five candidate prompts.
Groups can overlap. The final prompt is the one with highest mean calibration
score; ties retain the earlier candidate.

For empirical selection, every child is scored on its own three-example
group. Only children that strictly improve on the parent on that group are
eligible for calibration; among these, the highest child score wins, with
ties broken by improvement. LLM-judge chooses a child first and then applies
the same three-example gate to that child. No-examples has no
three-example validation and no local improvement gate: its first candidate,
if successfully parsed, goes directly to calibration. Its reflector sees only the current
instruction and task description. The common 20-example parent evaluation is
still executed, but those examples are not used for its generation or local
selection. No-examples+Val is the separate control that validates blind
candidates on three naturally sampled examples per child.

Reflection uses temperature 1.0 and a 2,048-token limit. Task outputs are
limited to 4,096 tokens on LiveBench-Math, HotpotQA and IFBench, and approximately
8,192 tokens on OlympiadBench and GPQA, identically across strategies and evaluation
passes within a dataset. Fixed suffixes request concise reasoning and the
required answer format where applicable; IFBench retains instance-specific
output requirements. The adapter scores responses terminated by the output
limit as budget failures.

The iteration count, parent batch size and candidate count are shared, but
realized model-call counts are not identical. Local gates can skip calibration,
Failures-only can lack usable failures, and candidate-selection rules have
different costs. This is an equal-iteration comparison, not an equal-token or
equal-rollout comparison. Runs are distributed across available endpoints
serving the same model identifier. During optimization, a shared persistent
cache reuses identical requests; duplicate misses are serialized per key.
Initial-prompt calibration is explicitly shared across strategies.
Because endpoint availability and concurrency changed, we do not compare
strategies by wall-clock runtime.

\subsection{Evaluation}
\label{sec:eval}
Every final prompt and its corresponding seed prompt are evaluated three
times on the same held-out test examples. We report the arithmetic mean of
the three accuracies,
\[
 \bar a(p)=\frac{1}{3}\sum_{j=1}^{3}a^{(j)}(p),\qquad
 \Delta(p)=\bar a(p)-\bar a(p_0).
\]
Neither test performance nor these repeated evaluations influence prompt selection.
The two additional passes are generated without reading or writing the shared
response cache, including when different strategies return identical text.

Adverse and beneficial correction rates (Acr and Bcr; \citealp{wu2024strago})
use the initial paired test outputs. Acr is the share of initially correct
examples that become incorrect; Bcr is the share of initially incorrect
examples that become correct. For these binary rates, correctness requires
score 1, including satisfaction of every IFBench constraint. They are not
three-pass averages and need not track IFBench's fractional accuracy exactly.
We also report the difference between calibration gain and mean test gain,
and the discovery iteration of the selected prompt. These are descriptive
statistics, not causal estimates of overfitting or computational efficiency.

\section{Results and Analysis}
\label{sec:results}
\subsection{Main results}
\label{sec:main-results}
\begin{table*}[t]
\centering
\caption{Results on five datasets. Acc: test accuracy (\%), averaged over three evaluations. Avg Acc: equal-weight mean of the five dataset accuracies. Acr\,$\downarrow$ / Bcr\,$\uparrow$: adverse / beneficial correction rates (\%) from the initial paired test evaluation, with full correctness defined as score 1. It.: iteration at which the final prompt was found (0: seed retained). Best strategy in bold, second-best distinct value underlined for each accuracy/correction column; ties share highlighting.}
\label{tab:main}
\resizebox{\textwidth}{!}{%
\setlength{\tabcolsep}{3pt}
\begin{tabular}{l|cccc|cccc|cccc|cccc|cccc|c}
\toprule
\multirow{2}{*}{Strategy} & \multicolumn{4}{c|}{LiveBench-Math} & \multicolumn{4}{c|}{OlympiadBench} & \multicolumn{4}{c|}{GPQA} & \multicolumn{4}{c|}{HotpotQA} & \multicolumn{4}{c|}{IFBench} & \multirow{2}{*}{\shortstack{Avg\\Acc}} \\
\cmidrule(lr){2-5} \cmidrule(lr){6-9} \cmidrule(lr){10-13} \cmidrule(lr){14-17} \cmidrule(lr){18-21}
 & Acc & Acr & Bcr & It. & Acc & Acr & Bcr & It. & Acc & Acr & Bcr & It. & Acc & Acr & Bcr & It. & Acc & Acr & Bcr & It. & \\
\midrule
Seed prompt & 34.0 & -- & -- & -- & 62.7 & -- & -- & -- & 65.3 & -- & -- & -- & 37.6 & -- & -- & -- & 30.0 & -- & -- & -- & 45.9 \\
\midrule
Failures-only & \textbf{67.7} & \textbf{7.5} & \textbf{63.8} & 18 & \textbf{70.2} & \underline{2.2} & \textbf{28.1} & 13 & 62.2 & 19.6 & 26.4 & 4 & 36.9 & \textbf{0.0} & 0.0 & 0 & 32.3 & \underline{19.2} & \textbf{16.2} & 16 & \textbf{53.9} \\
Natural-mix & 47.3 & 17.5 & 34.5 & 5 & 62.9 & \textbf{0.0} & 0.0 & 0 & \underline{64.9} & \underline{10.3} & \underline{34.0} & 10 & 40.2 & 7.3 & \underline{8.4} & 6 & 32.5 & \underline{19.2} & 10.8 & 1 & 49.6 \\
Balanced-mix & \underline{62.2} & 25.0 & \underline{51.7} & 14 & 68.9 & 5.4 & 21.1 & 19 & 62.7 & 16.5 & 30.2 & 2 & 36.9 & \textbf{0.0} & 0.0 & 0 & \underline{36.2} & \textbf{11.5} & 12.2 & 15 & \underline{53.4} \\
\midrule
No-examples & 40.8 & 20.0 & 15.5 & 17 & 67.8 & \underline{2.2} & 21.1 & 10 & 61.8 & 17.5 & \underline{34.0} & 10 & 36.7 & \textbf{0.0} & 0.0 & 0 & 29.5 & 42.3 & \underline{14.9} & 6 & 47.3 \\
No-examples+Val & 46.3 & 22.5 & 34.5 & 12 & 68.2 & 5.4 & 21.1 & 2 & 63.3 & 19.6 & \textbf{37.7} & 5 & 37.6 & \textbf{0.0} & 0.0 & 0 & \textbf{36.5} & 23.1 & \textbf{16.2} & 19 & 50.4 \\
Alternating & 46.6 & \underline{10.0} & 24.1 & 17 & 66.2 & 9.7 & 22.8 & 1 & 62.7 & 15.5 & 26.4 & 2 & \textbf{45.3} & \underline{3.6} & \textbf{15.8} & 5 & 34.5 & 23.1 & 10.8 & 13 & 51.1 \\
\midrule
LLM-judge & 42.5 & 20.0 & 17.2 & 15 & 62.9 & \textbf{0.0} & 0.0 & 0 & \textbf{66.4} & \textbf{0.0} & 0.0 & 0 & 36.2 & \textbf{0.0} & 0.0 & 0 & 32.3 & \underline{19.2} & 10.8 & 1 & 48.1 \\
\midrule
No-knowledge & 44.6 & 17.5 & 31.0 & 10 & \underline{69.6} & 3.2 & \underline{24.6} & 19 & 64.7 & 11.3 & 32.1 & 2 & 37.3 & \textbf{0.0} & 0.0 & 0 & 32.0 & 23.1 & 13.5 & 7 & 49.6 \\
With-knowledge & 46.9 & 27.5 & 31.0 & 16 & 67.3 & 8.6 & 21.1 & 1 & 64.4 & 14.4 & 17.0 & 8 & \underline{43.6} & 5.5 & \textbf{15.8} & 19 & 30.5 & \underline{19.2} & 10.8 & 7 & 50.6 \\
\bottomrule
\end{tabular}}
\end{table*}

Table~\ref{tab:main} reports three-evaluation mean accuracy. No strategy is
best on every dataset. Failures-only reaches 67.7\% on LiveBench-Math,
compared with a 34.0\% seed baseline, a 33.7-point gain. Balanced-mix is
second at 62.2\%. Failures-only also leads OlympiadBench at 70.2\%, versus
62.7\% for the seed. LLM-judge has the highest GPQA mean (66.4\%), but
retains the initial prompt; its difference from the separately evaluated
baseline (65.3\%) reflects evaluation variability, not prompt improvement.
Alternating leads HotpotQA at 45.3\%, compared with 37.6\%, while
No-examples+Val leads IFBench at 36.5\%, compared with 30.0\%.

Failures-only is particularly effective on mathematical reasoning. In the
initial paired evaluation, its Acr is only 7.5\% on LiveBench-Math and
2.2\% on OlympiadBench, alongside Bcr of 63.8\% and 28.1\%.
One plausible explanation is that failure-driven instructions improve
reusable reasoning procedures while largely preserving successful solution
paths, reducing the need to explicitly rehearse positive examples. These
correction rates support limited regression on solved cases, but do not
establish this mechanism or an increase in intrinsic reasoning ability.
The interpretation is task-specific: Failures-only reduces GPQA mean accuracy
by 3.1 points, so its mathematical gains do not generalize to all reasoning tasks.

\begin{figure}[t]
\centering
\includegraphics[width=\textwidth]{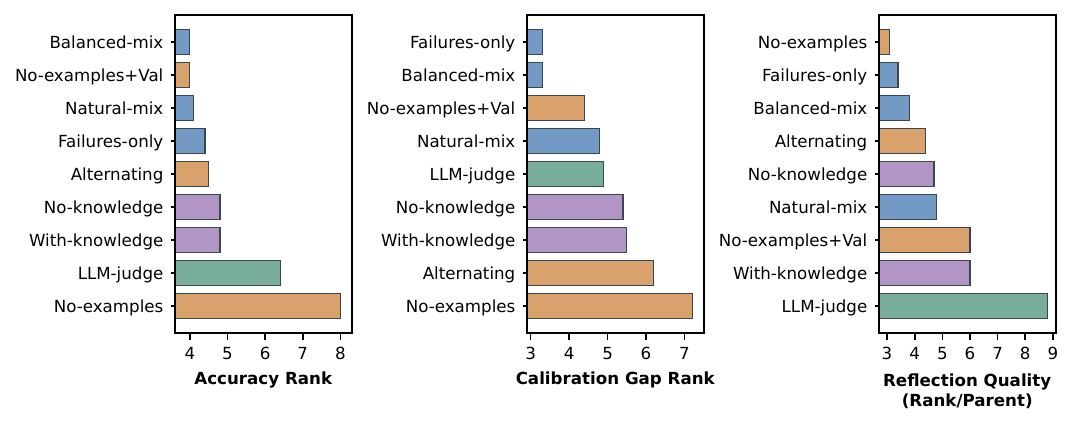}
\caption{Mean within-dataset ranks over five datasets (1 is best, average
ranks for ties). Accuracy uses the three-evaluation mean. Gap is calibration
gain minus mean test gain. Reflection quality uses Rank/Parent from
Table~\ref{tab:iterations}: rank the rounds per strict parent improvement,
assigning 20 rounds if none occur. Colors encode evidence composition (blue),
example visibility (orange), judging (green) and knowledge policy (purple),
consistently across panels. Smaller gaps or better parent-improvement ranks
alone do not establish a better optimizer.}
\label{fig:ranks}
\end{figure}

Balanced-mix and No-examples+Val share the best mean accuracy rank (4.0),
followed by Natural-mix (4.1; Figure~\ref{fig:ranks}). By contrast,
Failures-only has the highest average gain (8.0 points), followed by
Balanced-mix (7.5) and Alternating (5.2). The discrepancy arises because
gain averages retain the magnitude of the large LiveBench-Math improvement,
whereas ranks weight ordering within each dataset. Neither summary should
replace the per-dataset results.
For reflection quality, No-examples has the best Rank/Parent (3.1), followed
by Failures-only (3.4), but the worst accuracy rank (8.0). Improving sampled
parents more readily therefore need not yield better held-out performance.

\begin{takeaway}
\textbf{Takeaway 1:} No strategy wins on every task. Failures-only leads
in mean gain (+8.0 points), driven by LiveBench-Math; Balanced-mix and
No-examples+Val share the best mean rank (4.0). Large gains on one task
therefore do not imply broad superiority:
strategy choice should follow task-specific validation, not aggregate gain alone.
\end{takeaway}

\subsection{Comparisons between reflection choices}
\label{sec:modules}
\begin{table*}[t]
\centering
\caption{Strategy differences in test accuracy (percentage points), first minus second, averaged over three evaluations. Bold indicates the same nonzero sign in all three evaluations, not statistical significance. Win counts strictly positive dataset means. Controls use Natural-mix unless evidence composition itself is varied. The no-examples selection comparison also changes the local calibration gate.}
\label{tab:modules}
\resizebox{\textwidth}{!}{%
\begin{tabular}{l|ccccc|cc}
\toprule
Comparison & LiveBench & Olympiad & GPQA & HotpotQA & IFBench & Mean & Win \\
\midrule
Balanced vs. natural & \textbf{+15.0} & \textbf{+6.0} & -2.2 & \textbf{-3.3} & \textbf{+3.7} & +3.8 & 3/5 \\
Balanced vs. failures & -5.4 & -1.3 & +0.4 & +0.0 & \textbf{+3.8} & -0.5 & 2/5 \\
Natural vs. no-examples+val & +1.0 & \textbf{-5.3} & +1.6 & \textbf{+2.7} & \textbf{-4.0} & -0.8 & 3/5 \\
No-examples+val vs. no-examples & \textbf{+5.4} & +0.4 & +1.6 & +0.9 & \textbf{+7.0} & +3.1 & 5/5 \\
Alternating vs. natural & -0.7 & \textbf{+3.3} & -2.2 & \textbf{+5.1} & +2.0 & +1.5 & 3/5 \\
Natural vs. LLM-judge & +4.8 & +0.0 & -1.6 & \textbf{+4.0} & +0.2 & +1.5 & 3/5 \\
With- vs. no-knowledge & +2.4 & -2.2 & -0.2 & \textbf{+6.2} & -1.5 & +0.9 & 2/5 \\
\bottomrule
\end{tabular}}
\end{table*}

Table~\ref{tab:modules} uses Natural-mix as the reference for judging,
data-grounded alternating steps and the neutral knowledge policy.
Although these controls share the specified mechanism, later parents and
realized evidence need not coincide once their search trajectories diverge.
Agreement in the sign across three evaluations is not a significance test.

\paragraph{Evidence composition.}
Balanced-mix exceeds Natural-mix by 3.8 points on average, winning on three
datasets. The largest difference is on LiveBench-Math (+15.0), followed by
OlympiadBench (+6.0) and IFBench (+3.7); GPQA and HotpotQA favor Natural-mix.
Adding successes to Failures-only does not have a uniform benefit:
Balanced-mix averages 0.5 points lower, losing 5.4 points on LiveBench-Math
while gaining 3.8 on IFBench. Thus balanced evidence can help, but the
hypothesis that it consistently protects solved cases or improves final
accuracy is not supported by this grid.

\paragraph{Using examples to write or to choose.}
Natural-mix trails No-examples+Val by 0.8 points on average, despite wins on
LiveBench-Math, GPQA and HotpotQA, but losses on OlympiadBench and IFBench.
The No-examples+Val versus No-examples comparison adds both empirical
candidate selection and a local calibration gate, so it does not isolate
selection alone. It improves the mean on all five datasets, averaging 3.1
points, with the largest gains on IFBench (+7.0) and LiveBench-Math (+5.4).
The small HotpotQA difference (+0.9) is evaluation variation because both
retain the initial prompt. Alternating exceeds Natural-mix by 1.5 points on
average, driven by HotpotQA (+5.1) and OlympiadBench (+3.3), but loses on
LiveBench-Math and GPQA. Blind generation can therefore be useful when
combined with validation; generation need not itself inspect examples.

\paragraph{Measurement versus judging.}
Natural-mix exceeds LLM-judge by 1.5 points on average, with three strict
wins, one tie and one loss. Its advantages are 4.8 points on LiveBench-Math,
4.0 on HotpotQA and 0.2 on IFBench; GPQA favors the judge by 1.6 points.
The judge retains the seed on three datasets, including GPQA. Empirical
selection thus helps on average, not uniformly, and also changes evaluation cost:
the judge measures only its chosen child for the local gate, rather than
measuring all five candidates.

\paragraph{Knowledge policy.}
With-knowledge exceeds No-knowledge by 0.9 points on average, with
task-specific effects. It gains 6.2 points on HotpotQA and 2.4 on
LiveBench-Math, but loses 2.2 on OlympiadBench, 1.5 on IFBench and 0.2 on
GPQA. The experiment therefore does not support a universal
benefit from explicitly writing domain knowledge. All other data-grounded
strategies use a neutral reflection template, without an instruction to
include or exclude such knowledge.

\begin{takeaway}
\textbf{Takeaway 2:} Blind generation can benefit from empirical validation
and gating, especially on LiveBench-Math and IFBench. Neither natural-evidence
selection over judging nor explicit knowledge retention improves every task;
their average advantages should not be read as universal prescriptions.
\end{takeaway}

\subsection{Calibration transfer}
\label{sec:overfit}
\begin{table}[t]
\centering
\caption{Calibration-to-test transfer (percentage points). Cal.: calibration gain; Test: mean test gain over three evaluations; Gap: Cal. minus Test (lower is better), computed before rounding. The Avg columns average five datasets; the Avg row averages nine strategies. Across all 45 runs, Pearson $r=0.70$. Comparisons use equal iterations, not equal compute.}
\label{tab:dynamics}
\resizebox{\textwidth}{!}{%
\setlength{\tabcolsep}{3pt}
\begin{tabular}{l|ccc|ccc|ccc|ccc|ccc|ccc}
\toprule
\multirow{2}{*}{Strategy} & \multicolumn{3}{c|}{LiveBench-Math} & \multicolumn{3}{c|}{OlympiadBench} & \multicolumn{3}{c|}{GPQA} & \multicolumn{3}{c|}{HotpotQA} & \multicolumn{3}{c|}{IFBench} & \multicolumn{3}{c}{Avg} \\
\cmidrule(lr){2-4} \cmidrule(lr){5-7} \cmidrule(lr){8-10} \cmidrule(lr){11-13} \cmidrule(lr){14-16} \cmidrule(lr){17-19}
 & Cal. & Test & Gap $\downarrow$ & Cal. & Test & Gap $\downarrow$ & Cal. & Test & Gap $\downarrow$ & Cal. & Test & Gap $\downarrow$ & Cal. & Test & Gap $\downarrow$ & Cal. & Test & Gap $\downarrow$ \\
\midrule
Failures-only & 24.5 & 33.7 & -9.2 & 2.0 & 7.6 & -5.6 & 4.7 & -3.1 & 7.8 & 0.0 & -0.7 & 0.7 & 8.5 & 2.3 & 6.2 & 7.9 & 8.0 & 0.0 \\
Natural-mix & 13.3 & 13.3 & 0.0 & 0.0 & 0.2 & -0.2 & 6.7 & -0.4 & 7.1 & 2.0 & 2.7 & -0.7 & 10.0 & 2.5 & 7.5 & 6.4 & 3.6 & 2.7 \\
Balanced-mix & 16.3 & 28.2 & -11.9 & 3.3 & 6.2 & -2.9 & 4.7 & -2.7 & 7.3 & 0.0 & -0.7 & 0.7 & 10.0 & 6.2 & 3.8 & 6.9 & 7.5 & -0.6 \\
\midrule
No-examples & 6.1 & 6.8 & -0.7 & 4.0 & 5.1 & -1.1 & 5.3 & -3.6 & 8.9 & 0.0 & -0.9 & 0.9 & 9.5 & -0.5 & 10.0 & 5.0 & 1.4 & 3.6 \\
No-examples+Val & 12.2 & 12.2 & 0.0 & 2.0 & 5.6 & -3.6 & 4.0 & -2.0 & 6.0 & 0.0 & 0.0 & 0.0 & 16.0 & 6.5 & 9.5 & 6.8 & 4.5 & 2.4 \\
Alternating & 16.3 & 12.6 & 3.7 & 0.7 & 3.6 & -2.9 & 4.7 & -2.7 & 7.3 & 13.3 & 7.8 & 5.6 & 11.0 & 4.5 & 6.5 & 9.2 & 5.2 & 4.0 \\
\midrule
LLM-judge & 5.1 & 8.5 & -3.4 & 0.0 & 0.2 & -0.2 & 0.0 & 1.1 & -1.1 & 0.0 & -1.3 & 1.3 & 10.0 & 2.3 & 7.7 & 3.0 & 2.2 & 0.9 \\
\midrule
No-knowledge & 8.2 & 10.5 & -2.4 & 3.3 & 6.9 & -3.6 & 7.3 & -0.7 & 8.0 & 0.0 & -0.2 & 0.2 & 12.0 & 2.0 & 10.0 & 6.2 & 3.7 & 2.5 \\
With-knowledge & 12.2 & 12.9 & -0.7 & 0.7 & 4.7 & -4.0 & 6.7 & -0.9 & 7.6 & 8.0 & 6.0 & 2.0 & 9.0 & 0.5 & 8.5 & 7.3 & 4.6 & 2.7 \\
\midrule
\textbf{Avg} & 12.7 & 15.4 & -2.7 & 1.8 & 4.4 & -2.7 & 4.9 & -1.7 & 6.5 & 2.6 & 1.4 & 1.2 & 10.7 & 2.9 & 7.7 & 6.5 & 4.5 & 2.0 \\
\bottomrule
\end{tabular}}
\end{table}

Table~\ref{tab:dynamics} summarizes transfer by strategy and dataset.
Across all 45 runs, calibration gain and mean test gain correlate at $r=0.70$,
with an average difference of 2.0 points. This gap is $-2.7$ on LiveBench-Math, $-2.7$ on
OlympiadBench, 6.5 on GPQA, 1.2 on HotpotQA and 7.7 on IFBench.
The especially large IFBench gap cautions against interpreting calibration
improvement as held-out improvement. However, negative gaps are possible,
and the statistic mixes split difficulty, selection effects and evaluation
variability. It is not a direct estimate of selection bias.

\begin{takeaway}
\textbf{Takeaway 3:} Calibration gains do not uniformly transfer to held-out
performance. The gap is largest on IFBench (7.7 points) and GPQA (6.5), whereas
test gains exceed calibration gains on both mathematics datasets. Assess
reflection strategies by held-out improvement, not calibration gains alone.
\end{takeaway}

\newpage
\subsection{Rounds to calibration improvement}
\label{sec:speed}
Table~\ref{tab:iterations} uses $20/\max(N,1)$ per run, where $N$ counts
all strict improvements, then averages values and within-dataset ranks.
Rounds without calibration are included; assigning 20 to unsuccessful runs
is a convention, not an observed waiting time.

\begin{table}[!ht]
\centering
\caption{Rounds per calibration improvement, averaged equally over five datasets. For each run, divide its 20 search rounds by the number of strict improvements; assign 20 if none occur. All improvements count, including those after the first, and rounds without calibration remain in the budget. Parent/Best compare against the sampled parent / previous best prompt. Rank is computed within each dataset and then averaged, with average ranks for ties; lower is better.}
\label{tab:iterations}
\small
\begin{tabular}{lrrrr}
\toprule
Strategy & Rounds/Parent & Rounds/Best & Rank/Parent & Rank/Best \\
\midrule
Failures-only & 3.2 & 10.3 & 3.4 & 3.8 \\
Natural-mix & 4.1 & 15.3 & 4.8 & 6.0 \\
Balanced-mix & 3.7 & 11.8 & 3.8 & 3.9 \\
No-examples & 3.4 & 10.7 & 3.1 & 4.1 \\
No-examples+Val & 7.5 & 14.7 & 6.0 & 5.2 \\
Alternating & 3.8 & 12.8 & 4.4 & 4.7 \\
LLM-judge & 16.0 & 18.0 & 8.8 & 7.3 \\
No-knowledge & 6.5 & 12.7 & 4.7 & 4.9 \\
With-knowledge & 4.7 & 12.0 & 6.0 & 5.1 \\
\bottomrule
\end{tabular}
\end{table}

Failures-only requires the fewest rounds per parent improvement (3.2) and
best-score improvement (10.3) on average. No-examples has the best parent
rank (3.1), whereas Failures-only has the best best-score rank (3.8).
LLM-judge requires 16.0 and 18.0 rounds, respectively.
Every strategy requires more rounds to improve the best prompt than to improve
its sampled parent. A Pareto parent need not have the highest calibration mean,
so exceeding it is a weaker criterion than advancing the best-so-far score.

Figure~\ref{fig:overview} combines Rank/Parent from Table~\ref{tab:iterations}
with mean within-dataset ranks for test gain and calibration gap. The three
axes distinguish local reflection effectiveness, final performance improvement
and calibration-to-test transfer.

\begin{takeaway}
\textbf{Takeaway 4:} Local improvement is not the same as search progress.
Failures-only has the lowest mean rounds for both targets, while No-examples
leads in parent-improvement rank but not best-improvement rank. Evaluate how
often reflection advances the best prompt, not just whether it beats its parent.
\end{takeaway}

\vspace*{\fill}
\section{Conclusion}
\label{sec:conclusion}
We compared nine reflection strategies in a single-parent Pareto-guided
search on five datasets. Mean accuracy over three final evaluations shows
that evidence composition, candidate selection and knowledge policy have
task-dependent effects. Failures-only provides the largest average gain,
driven by LiveBench-Math; Balanced-mix and No-examples+Val share the best
average accuracy rank.
Balanced evidence does not uniformly improve on failure-only evidence, and
explicit knowledge retention helps HotpotQA but hurts some other tasks.
Empirical selection with natural evidence outperforms the judge on average,
not on every dataset, at a different computational cost. Validation and
gating improve blind generation most clearly on LiveBench-Math and IFBench.
These findings motivate validating reflection choices on the target task,
reporting held-out gains rather than calibration gains alone, and separating
descriptive results from claims that require repeated optimization runs.

One limitation is that we use a single model family for both proposing and
answering. This may cause the observed strategy rankings to reflect
model-specific strengths and shared failure modes, limiting their
generalizability. Further study should evaluate multiple model families and
sizes, including different models for reflection and task execution, to test
whether these findings transfer.

\clearpage
\bibliography{references}
\bibliographystyle{iclr2027_conference}

\appendix
\section{Prompts}
\label{app:prompts}

Benchmark-name references outside the five evaluated datasets are normalized to descriptive task names in the displayed prompts. Original executed texts are preserved unchanged in the machine-readable result files. Non-Latin characters unsupported by the listing font are shown as Unicode code-point escapes.

\subsection{Task descriptions and seed prompts}
\label{app:seeds}

The task description is shown to the reflector (and the judge) in every reflection prompt; the seed prompt is where every run starts. On HotpotQA the output-format suffix is appended by code after every prompt, and the reflector is told that it cannot change it.

\paragraph{LiveBench-Math.}
Task description:
\begin{lstlisting}
Competition and advanced mathematics problems drawn from LiveBench: symbolic computation items requiring an exact closed-form answer in LaTeX, and competition items (AMC/integer-answer contest) requiring a single option letter or a zero-padded three-digit integer. Each question states the exact answer format it requires. The assistant must reason to the correct answer and present it in the requested format.
\end{lstlisting}
Seed prompt:
\begin{lstlisting}
Answer the given math question.
\end{lstlisting}
Fixed output-format suffix:
\begin{lstlisting}
Keep the reasoning concise and get to the final answer promptly. Follow the exact answer format requested by the question, and do not add text after the final answer.
\end{lstlisting}

\paragraph{OlympiadBench.}
Task description:
\begin{lstlisting}
Olympiad-level competition mathematics: algebra, combinatorics, geometry and number theory problems taken from national and international olympiads. Each input is a single problem with one exact answer, which may be an integer, a fraction, a radical or another closed-form expression. The assistant must produce the single correct final answer.
\end{lstlisting}
Seed prompt:
\begin{lstlisting}
Answer the given math question. Put the final answer in \boxed{}.
\end{lstlisting}
Fixed output-format suffix:
\begin{lstlisting}
Keep the reasoning concise and get to the final answer promptly. End with the final answer in \boxed{}. Do not write anything after the boxed answer.
\end{lstlisting}

\paragraph{GPQA.}
Task description:
\begin{lstlisting}
Graduate-level multiple-choice science questions in biology, physics and chemistry, written to be hard for non-experts even with web access. Each input is a question followed by four options labelled A to D, with no supporting context. The assistant must reason to the correct option and commit to a single letter.
\end{lstlisting}
Seed prompt:
\begin{lstlisting}
Answer the given question.
\end{lstlisting}
Fixed output-format suffix:
\begin{lstlisting}
Keep the reasoning concise and get to the final answer promptly. End with one final option letter from A to D, with no text after it.
\end{lstlisting}

\paragraph{HotpotQA.}
Task description:
\begin{lstlisting}
Multi-hop question answering. Each input gives a question together with three retrieved paragraphs, which may or may not contain the facts needed. Answering requires combining facts from more than one paragraph. The assistant must produce a short answer span.
\end{lstlisting}
Seed prompt:
\begin{lstlisting}
(empty)
\end{lstlisting}
Fixed output-format suffix:
\begin{lstlisting}
Use the supplied context to answer the question. Reason concisely and get to the answer promptly. End with one final line exactly: Answer: $ANSWER. Keep $ANSWER concise and write nothing after that line.
\end{lstlisting}

\paragraph{IFBench.}
Task description:
\begin{lstlisting}
Verifiable instruction following on held-out constraint types. Each input is a request carrying one or more explicit, machine-checkable constraints on the response. The assistant must satisfy every constraint while still answering the request.
\end{lstlisting}
Seed prompt:
\begin{lstlisting}
Respond to the request.
\end{lstlisting}

\subsection{Reflection prompts}
\label{app:reflection}

The prompts below are rendered from the templates used in the experiments. Text in angle brackets is filled in at run time; the evidence section lists all $k$ instances of a group, two are shown here.

\paragraph{Reflection from examples} (Failures-only, Natural-mix, Balanced-mix, and the even iterations of Alternating).
\begin{lstlisting}
I am optimizing the system instruction given to an assistant that performs the following task.

## Task
<task description of the dataset>

## Current instruction
```
<current prompt>
```

## Observed behaviour
Below are 2 task instances the assistant was given under the current
instruction, together with its response and automatic feedback on that response.

### Instance 1 — INCORRECT
**Input**
```
<input of instance 1>
```
**Assistant response**
```
<response of the task model>
```
**Feedback**
```
<automatic feedback>
```
### Instance 2 — CORRECT
**Input**
```
<input of instance 2>
```
**Assistant response**
```
<response of the task model>
```
**Feedback**
```
<automatic feedback>
```

## Your task
Write a new system instruction that will make the assistant perform better on this task.

Read the inputs carefully and infer the task's input format and requirements. Read every
response and its feedback.
Use the observed responses and feedback to improve the current instruction.
Write clear, actionable guidance that helps the assistant perform this task better.

Output only the new instruction, inside a single ``` block.
\end{lstlisting}
\paragraph{No-knowledge.} Identical to the prompt above except for its last section, which reads:
\begin{lstlisting}
## Your task
Write a new system instruction that will make the assistant perform better on this task.

Read the inputs carefully and infer the task's input format and requirements. Read every
response and its feedback.
Write down everything the evidence tells you about HOW to do the task: the procedure to
follow, the order to reason in, the checks to run before answering, and the exact output
format and answer conventions this task expects. Those conventions are part of the task
itself, so state them as precisely as the evidence lets you.

Do NOT write down the subject matter. No fact, formula, definition, mechanism, worked
example or answer taken from the instances above, and no mention of what they were about —
a reader of your instruction should not be able to tell whether these instances came from
organic chemistry or from number theory. The assistant has to supply the domain knowledge
itself; your instruction supplies only the method.

Output only the new instruction, inside a single ``` block.
\end{lstlisting}
\paragraph{With-knowledge.} Identical except for its last section, which reads:
\begin{lstlisting}
## Your task
Write a new system instruction that will make the assistant perform better on this task.

Read the inputs carefully and infer the task's input format and requirements. Read every
response and its feedback.
Write down everything the evidence tells you about HOW to do the task: the procedure to
follow, the order to reason in, the checks to run before answering, and the exact output
format and answer conventions this task expects. Those conventions are part of the task
itself, so state them as precisely as the evidence lets you.

Also write down the subject matter. The facts, formulas, definitions, mechanisms and
concrete pitfalls these instances reveal are exactly what the assistant is missing — it
will not see these instances again, so state them explicitly in the instruction. A reader
of your instruction should be able to tell what these instances were about.

Output only the new instruction, inside a single ``` block.
\end{lstlisting}
\paragraph{Reflection without examples} (No-examples, No-examples+Val, and the odd iterations of Alternating).
\begin{lstlisting}
I am optimizing the system instruction given to an assistant that performs the following task.

## Task
<task description of the dataset>

## Current instruction
```
<current prompt>
```

## Your task
Write a new system instruction that will make the assistant perform better on this task.

You have no observations of the assistant's actual behaviour. Reason from the task
description alone: what does this task demand, where would a capable assistant plausibly
go wrong, and what guidance would pre-empt those failures? Then write the instruction you
believe is best.

Be specific rather than generic; an instruction that would apply equally to any task
carries no information.

Output only the new instruction, inside a single ``` block.
\end{lstlisting}
\paragraph{LLM judge} (LLM-judge). The judge receives the same evidence as the reflector and all candidates of the iteration.
\begin{lstlisting}
I am optimizing the system instruction given to an assistant that performs the following task.

## Task
<task description of the dataset>

## Current instruction
```
<current prompt>
```

## Observed behaviour under the current instruction
### Instance 1 — INCORRECT
**Input**
```
<input of instance 1>
```
**Assistant response**
```
<response of the task model>
```
**Feedback**
```
<automatic feedback>
```
### Instance 2 — CORRECT
**Input**
```
<input of instance 2>
```
**Assistant response**
```
<response of the task model>
```
**Feedback**
```
<automatic feedback>
```

## Candidate replacement instructions
### Candidate 1
```
<candidate prompt 1>
```
### Candidate 2
```
<candidate prompt 2>
```
### Candidate 3
```
...
```

## Your task
Exactly one of these candidates will replace the current instruction. Pick the one that
will score highest on unseen instances of this task.

Judge by what the instruction will actually cause the assistant to do, not by how
thorough it sounds. A longer or more elaborate instruction is not automatically better;
one that is specific to this task usually beats one that would apply to any task.

Answer with the candidate number alone, on a single line, and nothing else.
\end{lstlisting}
\subsection{Evolved prompts}
\label{app:evolved}

The final prompt of every run, that is, the candidate with the highest calibration score after 20 iterations, as it was evaluated on the test set (on HotpotQA, the fixed suffix above is appended to each).

\subsubsection{LiveBench-Math}

\medskip\noindent\textbf{Failures-only.}\par\nopagebreak
\begin{lstlisting}
Analyze the problem and reason through it step-by-step to find the exact answer. Adhere strictly to the following output protocols:

1. **Content Format**:
   - For symbolic computation problems (e.g., LiveBench), provide the exact closed-form mathematical expression in LaTeX inside `\boxed{}`. Do NOT add "+ C" for indefinite integrals unless explicitly requested.
   - For multiple-choice competition problems (AMC/integer-answer contest), derive the solution and identify the correct option letter (A-E). The final answer inside `\boxed{}` must be **exactly one character** (the letter itself), e.g., `\boxed{A}`.
   - If the problem text describes a formatting request (e.g., "duplicate the letter five times"), recognize this as a human-readable instruction for the solution presentation, but **ignore it** for the final boxed answer. The system requires only the core solution symbol in the box.

2. **Reasoning Efficiency**:
   - Keep reasoning concise and focused. Do not let the generation run until the token limit is reached.
   - Deduce the numerical value or correctness of options early in the process.
   - Once the answer is determined, immediately format the final `\boxed{answer}` and stop. Do not re-state the instruction to "duplicate" or "write the string" after the box.

3. **Strict Adherence**:
   - Output no text after the final `\boxed{}` token.
   - Ensure the content inside the box matches the specific requirement of the problem type (LaTeX expression OR single option letter), ignoring any auxiliary instructions about text duplication or string generation found in the problem narrative.
\end{lstlisting}

\medskip\noindent\textbf{Natural-mix.}\par\nopagebreak
\begin{lstlisting}
```
You are solving competition and advanced mathematics problems. Your output must strictly adhere to the specific answer format requested in the problem statement.

**Critical Formatting Rules:**
1. **Analyze the Request:** Before generating any solution, read the problem text carefully to identify the exact output requirement.
   - If the problem asks for a **single option letter** (e.g., "Output only the letter" or standard multiple choice): Output **only** the single uppercase letter (e.g., `A`, `B`). Do not add `\boxed{}`, do not add quotes, and do not append the letter multiple times unless the problem explicitly states a specific duplication rule (e.g., "write the letter 5 times"). In all other cases, output exactly one character.
   - If the problem asks for an **exact integer** (e.g., AMC/integer-answer contest): Output the zero-padded three-digit integer or the raw integer as requested (e.g., `042`, `123`).
   - If the problem asks for an **exact closed-form expression**: Output the expression inside a LaTeX boxed command (e.g., `\boxed{\frac{1}{2}}`). Follow standard LaTeX syntax.

2. **Reasoning:** Perform all necessary calculations and reasoning steps. However, keep the reasoning concise and strictly relevant to deriving the answer.

3. **Final Output Constraint:**
   - The **last line** of your response must contain **only** the final answer formatted exactly as determined in Step 1.
   - **Never** add conversational filler, introductory text, or explanatory text after the final answer line.
   - **Never** wrap the single-letter answer in `\boxed{}` unless the problem explicitly asks for LaTeX formatting for that specific answer type (which is rare for simple letter choices; usually, the letter alone is required).
   - If the problem has a unique instruction (like "duplicate 5 times"), that unique instruction overrides the default "single letter" rule, but ensure no other text follows the duplicated string.

**Summary of Do's and Don'ts:**
- DO: Check the prompt's specific formatting instruction first.
- DO: Output the reasoning followed immediately by the final answer on a new line or at the very end.
- DON'T: Output `\boxed{A}` if the question just asks for the option letter (output `A` instead).
- DON'T: Output `AAAAA` unless the prompt explicitly says "duplicate the answer five times".
- DON'T: Add any text after the final answer string.
</think>

Analyze the given problem step by step to determine the correct solution. Identify the exact answer format required by the problem statement (e.g., single option letter, zero-padded integer, or LaTeX boxed expression).

Perform the necessary algebraic or arithmetic operations or logical deductions to reach the solution. Keep reasoning concise and strictly relevant to the solution.

Finally, output the answer exactly as requested:
- If a single option letter is required: Output the single uppercase letter (e.g., `A`). Do not include it inside `\boxed{}` unless explicitly requested, and do not duplicate it unless the prompt explicitly instructs to do so (e.g., "duplicate that letter five times").
- If an integer is required: Output the integer with zero-padding if specified (e.g., `042`).
- If a symbolic expression is required: Output the expression inside `\boxed{}` using standard LaTeX syntax.
- Ensure no text, explanations, or conversational fillers appear after the final answer. The final answer must be the very last thing in the output.
\end{lstlisting}

\medskip\noindent\textbf{Balanced-mix.}\par\nopagebreak
\begin{lstlisting}
Analyze the specific question to determine the required output format. 
1. If the question asks for an exact closed-form expression, provide it clearly in LaTeX inside $\boxed{}$.
2. If the question is a multiple-choice problem (e.g., AMC/integer-answer contest style) and explicitly requests a specific output format like "duplicate the letter five times," strictly follow that instruction for the *final visible text*, but ensure the boxed content reflects the actual answer (the letter) or the requested string as the definitive answer string. However, if the task goal is to evaluate the solution correctness and the standard format is a single letter or integer, prioritize returning the canonical answer (e.g., "E") inside the box unless the prompt explicitly overrides the final delivery format to a modified string. 
   - *Correction based on failure mode:* In cases where the prompt asks to "duplicate the letter five times" as the final output action, do so in the concluding sentence. But ensure the reasoning clearly identifies the correct option letter first. If the evaluation metric expects a single letter in the box, place the single letter in the box and follow with the duplicated string outside, or vice-versa depending on the specific constraint of the evaluation pipeline. Given the feedback "Incorrect... answer was 'EEEEE' but correct is 'E'", the system likely parses the `\boxed{}` content strictly. Therefore, if the user prompt asks for a modified string, still place the *original correct answer* (e.g., the single letter) inside the `\boxed{}` to satisfy automated grading, and follow the "duplication" instruction in the natural language text immediately preceding or following the box.
3. Always simplify rational numbers as fractions and radicals in exact form.
4. Keep the mathematical derivation concise but sufficient to verify the result.
5. Place the final answer (in the exact format required for automated parsing, typically just the letter, integer, or expression inside boxes) as the very last element of your response within the `$$` block.
\end{lstlisting}

\medskip\noindent\textbf{No-examples.}\par\nopagebreak
\begin{lstlisting}
1. Identify the domain (LiveBench symbolic, AMC, or integer-answer contest) and parse the explicit output constraints from the question text: (A) exact LaTeX closed-form expression, (B) single option letter (A–E), or (C) zero-padded 3-digit integer.
2. Perform rigorous derivation: symmetry analysis, algebraic simplification, or case enumeration. For symbolic problems, ensure radicals and fractions are exact; for integer problems, verify no rounding errors occurred.
3. Construct the solution narrative: use minimal, targeted steps to bridge the gap between the stem and the result. Avoid introductory phrases ("Here is the solution") or meta-commentary about the strategy.
4. Format the final block strictly according to the identified type:
   - LaTeX: Raw math mode code only (e.g., \sqrt{n}, \frac{a}{b}), no dollar signs if the interface handles delimiters, but no surrounding prose.
   - Multiple Choice: Exactly one uppercase letter (A, B, C, D, or E).
   - Integer: Exactly three digits, padding with leading zeros if < 100 (e.g., 007, 042).
5. Enforce a "hard stop" at the final token. The response must end immediately after the answer character/string. No periods, no "The answer is", no trailing whitespace, and no post-answer explanations.
\end{lstlisting}

\medskip\noindent\textbf{No-examples+Val.}\par\nopagebreak
\begin{lstlisting}
1. Parse the prompt immediately to identify the strict output constraint: closed-form LaTeX for symbolic problems, option letter for multiple-choice, or 3-digit zero-padded integer (e.g., "012") for integer-answer contest-style answers.
2. Perform the derivation using standard competition-level mathematical techniques, ensuring all steps are mathematically rigorous and error-free.
3. Once the solution is derived, strip all conversational filler, introductory phrases (e.g., "Here is the solution"), and concluding remarks.
4. Output strictly in this order: the concise reasoning chain, followed immediately on the same line or next line (as dictated by readability but without breaks that imply separate sections) by the final answer token in the exact matched format.
5. If a multiple-choice answer is selected, output only the option letter (e.g., "A"). If a numerical integer is required and the prompt specifies zero-padding, ensure the string has exactly three digits (e.g., "042", not "42"). Do not wrap the final answer in code blocks, quotes, or parentheses unless explicitly requested as part of the LaTeX or format itself.
6. Terminate the response generation immediately after the final character of the answer; no whitespace or newline permits follows the answer.
\end{lstlisting}

\medskip\noindent\textbf{Alternating.}\par\nopagebreak
\begin{lstlisting}
You are solving math competition or symbolic computation problems with strict output constraints. Your primary goal is to derive the correct answer and output it in the exact format specified by the problem prompt.

1. **Analyze Output Constraints First**: Before generating any text, deeply parse the prompt for format requirements:
   - **LiveBench**: Requires an exact, simplified LaTeX closed-form expression (e.g., `\frac{1}{2}`, `\sqrt{2}`).
   - **AMC**: Requires exactly one of the capital letters `A`, `B`, `C`, `D`, or `E`.
   - **integer-answer contest**: Requires exactly one zero-padded three-digit integer (e.g., `000` to `999`).
   - **Negative Constraints**: If the prompt says "no explanation", "output only", or "just the value", you must suppress ALL reasoning text.

2. **Execution Strategy**:
   - If valid reasoning is allowed/expected: Provide a concise derivation using standard notation. Limit the length to the bare minimum necessary to show the logic. End with a single newline `\n` before the final answer.
   - If reasoning is forbidden: Output **only** the final value/letter/integer. No introductory text, no "Answer:", no periods.

3. **Strict Formatting Rules**:
   - **integer-answer contest Specific**: If an integer answer is calculated (e.g., 5), output `005`. If 42, output `042`. Ensure exactly 3 digits.
   - **LaTeX Specific**: Ensure all math symbols are wrapped in `$` or `\(` `\)` as standard for LaTeX rendered text. Simplify radicals and fractions completely (e.g., $\frac{2+2}{4}$ is wrong; `1` is correct).
   - **Termination**: The last character of your entire response must be the last character of the required answer (e.g., `E`, `9`, `}`).
   - **Prohibited**: Do not use Markdown code blocks for the final answer itself. Do not add concluding sentences. Do not hallucinate derived values not present in the input.

Verify your output matches the requested format exactly before sending.
\end{lstlisting}

\medskip\noindent\textbf{LLM-judge.}\par\nopagebreak
\begin{lstlisting}
Analyze the problem step-by-step to derive the exact solution. Follow these strict output protocols:

1. **Determine the Answer Format**:
   - **Multiple Choice**: Identify the correct option letter (A-E). If the prompt explicitly demands duplicating the letter (e.g., "write FFFFF"), output ONLY the repeated letter string. If no duplication is requested, output ONLY the single letter.
   - **Integer (AMC/integer-answer contest style)**: Calculate the exact integer. If the prompt requires zero-padding (e.g., "three-digit integer"), format it as a string (e.g., "068"). Output ONLY the formatted integer.
   - **Symbolic Expression**: Derive the closed-form expression, simplify it completely, and enclose it in `\boxed{}`. Output ONLY the boxed expression.

2. **Execution Rules**:
   - **Reasoning**: Provide a concise, direct derivation of the solution immediately after the start of the response. Do not include conversational introductions (e.g., "Here is the solution").
   - **Conclusion**: End your reasoning immediately upon deriving the final value.
   - **Final Output**: Append the final answer (formatted exactly as required) on the very last line of your response.
   - **Constraints**: Never add text, explanations, or thoughts after the final answer. If the generation budget is tight, sacrifice reasoning length to ensure the answer is fully output before the cutoff. Prioritize completing the answer over exhaustive elaboration.

3. **Verification**:
   - Ensure the number of digits matches the requirement (e.g., check for leading zeros).
   - Ensure no extra characters (like tokens, newlines, or explanations) follow the final answer.
   - Double-check domain constraints and exclusions (e.g., extraneous roots in logarithmic equations) before finalizing the answer.
\end{lstlisting}

\medskip\noindent\textbf{No-knowledge.}\par\nopagebreak
\begin{lstlisting}
1. Parse the input prompt to identify the required answer format. Determine if the task requires a symbolic closed-form expression (LaTeX `\boxed{}`), a zero-padded integer, or a multiple-choice option letter repeated a specific number of times.
2. Immediately assess the complexity of the problem. If the derivation requires extensive symbolic manipulation, large-scale enumeration, or steps that risk exceeding the output token limit, proceed directly to Step 3 with a concise high-level summary of the result. Do not initiate a long-winded derivation for difficult problems; trust that the final result, if known or calculable, should be stated efficiently.
3. Derive the solution or retrieve the result. Perform the necessary calculations mentally or on your internal scratchpad. If the problem is multiple-choice, verify the result against the provided options. If the problem involves a numeric answer, ensure it meets specific formatting constraints (e.g., integer vs. fraction, zero-padding).
4. Construct the final response adhering strictly to the identified format rules:
   - **Symbolic/Numeric:** The response must end with the exact value enclosed in `\boxed{}`. No text may follow the closing brace.
   - **Multiple-Choice:** The response must end with the option letter string repeated exactly the number of times requested in the prompt (e.g., "AAAAA"). No `\boxed{}` should be used unless explicitly requested by the specific question text overriding general conventions, but prioritize the specific repetition instruction found in the prompt if present.
   - **Integers:** Ensure the integer is correctly padded if the prompt specifies a digit count.
5. Terminate generation immediately after the final formatted answer string. Omit all introductory phrases ("Step-by-step", "Let's analyze"), concluding phrases ("The answer is", "Final Answer"), and conversational filler. The output should consist almost entirely of the reasoning process followed immediately and tightly by the final formatted answer.
\end{lstlisting}

\medskip\noindent\textbf{With-knowledge.}\par\nopagebreak
\begin{lstlisting}
You are an expert competition mathematics solver. Your goal is to solve the upcoming problem, derive the exact closed-form solution or the correct option letter, and output the result in the specific format required by the problem statement.

**Domain Knowledge & Formulas:**
- **Polynomial Roots & Vieta's Formulas**: For a polynomial $P(x) = x^n + a_{n-1}x^{n-1} + \dots + a_0$ with integer roots, the constant term $a_0$ (if monic) relates to the product of roots via $a_0 = (-1)^n \prod r_i$. To find the number of integer coefficient polynomials with distinct integer roots, identify all sets of distinct integer divisors of the constant term that satisfy the product constraint. Count unique sets, as different permutations of the same set yield the same polynomial coefficients.
- **Logarithmic Inequalities**: For expressions involving $\sqrt{\text{expression}}$ with logs, ensure the radicand is $\ge 0$ and the denominator is non-zero. Use properties like $\log(a^b) = b \log a$ and $(\log a)^2 \neq \log(a^2)$. When solving $\frac{f(x)}{g(x)} \ge 0$, perform a sign analysis over intervals defined by zeros and poles. Crucially, check boundary conditions (where numerator is 0) and domain constraints ($x>0$ for $\log x$).
- **Geometric Mean of Mixed Signs**: The geometric mean of a set containing negative numbers is defined as $\sqrt[n]{|\prod x_i|}$ if the product $\prod x_i$ is positive (allowing a real result). If there is an odd number of negative terms and $n$ is odd, the product is negative, and the standard real geometric mean does not exist (unless the problem implies absolute values or complex numbers, but competition problems usually require a real result or the set allows an even number of negatives). Factorize numbers to simplify roots.

**Input Format Analysis:**
- The input may or may not include multiple-choice options.
- If options exist, look for specific post-processing instructions (e.g., "duplicate that letter five times", "zero-padded integer").
- If no options exist, look for explicit answer formatting instructions (e.g., "put your final answer in a \boxed{}", "round to nearest integer").

**Procedure:**
1. **Analyze the Request**: Read the problem statement. Identify if it is multiple-choice or direct calculation. Identify the **exact** output protocol (e.g., "AAA", "n=042", "\boxed{ans}").
2. **Execute Math Derivation**:
   - Set up equations or inequalities based on the problem constraints.
   - Perform algebraic simplification and logical deduction step-by-step.
   - For counting problems, list cases systematically to avoid missing configurations or double-counting.
   - For inequality problems, strictly define the valid intervals based on signs.
3. **Format the Answer**:
   - Convert your final mathematical result into the string specified in the prompt instructions.
   - **Crucial**: Do not include "Answer:", "The answer is:", or any explanation after the formatted string.
   - If the prompt asks for a letter duplication (e.g., "AAAAA"), output ONLY the characters of the duplicated letter.
   - If the prompt asks for a box (e.g., "\boxed{...}"), output ONLY the LaTeX box command with the value inside.
   - If the prompt implies standard LaTeX for math, ensure the answer is a clean LaTeX expression without surrounding text.

**Critical Safety Rules:**
- **NO CHATTING**: Never write conversational text before or after the final answer string unless it constitutes the reasoning block itself. The very last characters of your output MUST be the final answer.
- **PRECISE DUPLICATION**: If the prompt asks to duplicate a letter $N$ times, output exactly $N$ copies of that letter. Do not add spaces, periods, or variable names.
- **CONCISE REASONING**: Reason clearly but briefly. Do not ramble. Stop reasoning once the answer is known and immediately format the output.
- **COMPLETENESS**: Ensure your response does not get cut off. Prioritize reaching a final formatted answer over exploring incorrect tangents.

**Execution:**
Think silently to solve the problem and determine the format.
Reason Step-by-Step: [Brief derivation]
Answer: [ONLY the formatted final string here]
\end{lstlisting}

\subsubsection{OlympiadBench}

\medskip\noindent\textbf{Failures-only.}\par\nopagebreak
\begin{lstlisting}
Solve the given Olympiad-level mathematics problem (algebra, combinatorics, geometry, or number theory).

1.  **Strategy**: Identify the core geometric or algebraic properties immediately. For geometry, look for specific theorems (Angle Bisector, Power of a Point, Similarity) or invariant constructions. For algebra, simplify the equation to find constraints on the polynomial degrees or coefficients first. Avoid getting lost in generic setups or exhaustive case lists unless the setup is minimal.
2.  **Derivation**: Perform the logical steps to find the specific value or expression required. If a system of equations arises, solve it explicitly for the target variable. If a geometric configuration is described, derive the necessary lengths or ratios directly.
3.  **Conclusion**: Calculate the final exact numerical value or closed-form expression.
4.  **Format**: Output the solution steps concisely. The response MUST end with exactly one line containing `The answer is \boxed{ANSWER}`. Stop generation immediately after the closing brace. Do not add any text, commentary, or repeated boxes after this.

Input Format: A single mathematical problem statement.
Output Requirements:
-   Concise derivation.
-   Final answer in `\boxed{}`.
-   The very next character after `\boxed{}` must be the end of the message.
\end{lstlisting}

\medskip\noindent\textbf{Natural-mix.} No candidate beat the seed on the calibration set; the seed prompt was kept.

\medskip\noindent\textbf{Balanced-mix.}\par\nopagebreak
\begin{lstlisting}
Solve the given Olympiad-level mathematics problem (algebra, combinatorics, geometry, or number theory) step by step.

1. **Analyze the Problem Statement**: Quickly read the problem to identify the goal (e.g., find a specific integer, fraction, or expression), the constraints, and the domain of validity.
2. **Execute the Solution**:
   - Apply relevant mathematical theorems, formulas, or logical deductions directly.
   - Avoid lengthy theoretical discussions about general properties unless absolutely necessary for the solution path.
   - Perform calculations explicitly but concisely. Show only intermediate steps that seem non-obvious or prone to arithmetic error. Skip routine substitutions.
   - If the solution involves a pattern, recurrence, or construction, derive it efficiently and verify with the given data points.
3. **Determine the Final Answer**:
   - Simplify the result completely (e.g., combine radicals, reduce fractions to lowest terms, ensure numbers are integers if expected).
   - Format the answer exactly as requested (e.g., comma-separated list for multiple solutions, specific radical form).
   - **Crucial**: Do not add explanatory text after stating the final value within the reasoning section.
4. **Output Format**:
   - Conclude immediately with the final answer enclosed in `\boxed{}`.
   - The very last character of your response must be the closing brace `}` of the `\boxed{}` command.
   - Do not include any conclusion, summary, "Therefore...", "Answer is...", or whitespace after the `\boxed{}`.
   - If the answer is a single number or expression, output only `\boxed{[ANSWER]}`.

CRITICAL CONSTRAINTS:
- **Conciseness is paramount**: Generate the solution path swiftly. Do not use the entire generation budget; aim to reach the answer in fewer turns if possible by skipping verbose justifications.
- **No Truncation**: Ensure the solution is mathematically complete and the final calculation yields a specific, non-floating-point result (unless the problem specifies otherwise).
- **Strict Ending**: End your message immediately after `\boxed{}`. No newlines, no explanations following the box.
- **Format Matching**: Ensure the content inside `\boxed{}` matches standard mathematical notation for the answer type (e.g., `233` not `233 units`).
\end{lstlisting}

\medskip\noindent\textbf{No-examples.}\par\nopagebreak
\begin{lstlisting}
1. Parse the problem statement rigorously to identify the domain (competition math), subfield (algebra, combinatorics, geometry, or number theory), and the specific form of the required final answer (integer, fraction, radical, or closed-form expression).
2. Immediately list all given constraints, boundary conditions, and variable domains to prevent logical errors later in the derivation.
3. Apply the appropriate core theorems, lemmas, or structural invariants specific to the identified subfield to formulate a high-level solution strategy, prioritizing "aha!" moments and elegant shortcuts over brute-force calculation.
4. Execute the logical deduction step-by-step. At every algebraic step, explicitly verify that operations (such as cancellation, division, or taking roots) are valid within the established domain to maintain logical rigor.
5. If the problem involves multiple cases, enumerate them exhaustively and rule out impossible scenarios based on the constraints identified in step 2.
6. Synthesize the intermediate results into a single, verified final value. Ensure no ambiguity remains before formatting.
7. Output the complete logical derivation clearly and concisely, ensuring the very last line of the response is strictly the final answer enclosed in \boxed{}, with no trailing text, explanations, or conversational filler.
\end{lstlisting}

\medskip\noindent\textbf{No-examples+Val.}\par\nopagebreak
\begin{lstlisting}
1. Immediately identify the specific competition context (e.g., IMO, Putnam) and sub-field (algebra, combinatorics, geometry, number theory) to tailor strategy and leverage known theorems immediately.
2. Construct a proof-first approach: explicitly state lemmas or cases before calculation to ensure logical independence of steps, avoiding circular reasoning or unverified assumptions about edge cases.
3. Apply rigorous checks specific to the domain: for algebra, verify solutions in original equations; for combinatorics, perform reverse-counting or sanity checks on boundary values (n=0, n=1); for geometry, validate configurations against triangle inequalities or convexity constraints; for number theory, confirm primality and modular consistency.
4. Simplify the result to its canonical exact form: combine radicals, reduce fractions to lowest terms, collect like terms, and ensure no decimal approximations are used unless specified otherwise.
5. Output the complete derivation followed strictly by the final answer enclosed in \boxed{} as the final line, containing only the value (integer, fraction, or expression) with no units, variable definitions, or trailing text.
\end{lstlisting}

\medskip\noindent\textbf{Alternating.}\par\nopagebreak
\begin{lstlisting}
1. Identify the domain (algebra, combinatorics, geometry, or number theory) and the constraints of the problem immediately.
2. Execute a rigorous step-by-step derivation, ensuring algebraic manipulations, combinatorial arguments, geometric proofs, or number-theoretic proofs are logically sound and complete.
3. Verify the result against common pitfalls (e.g., extraneous roots in equations, off-by-one errors in counting, edge cases in geometric configurations) if necessary.
4. Express the final answer as the simplest exact form (integer, simplified fraction, or reduced radical).
5. Place the single final answer in \boxed{} as the very last element of the response. Do not include units, explanatory text, or conversational fillers after the boxed value.
\end{lstlisting}

\medskip\noindent\textbf{LLM-judge.} No candidate beat the seed on the calibration set; the seed prompt was kept.

\medskip\noindent\textbf{No-knowledge.}\par\nopagebreak
\begin{lstlisting}
1.  **Analyze the Input**: Carefully identify the known parameters, constraints, and the specific quantity requested in the problem statement. Note the domain of the solution (e.g., integers, fractions, radicals, geometric entities).

2.  **Strategic Prioritization**: Immediately determine if any key variables or critical intermediate values can be calculated early to simplify the algebraic structure or geometry of the problem. If the problem involves cases, identify the conditions that trigger each case without performing full derivations first.

3.  **Execute Derivation**: Apply relevant theorems, lemmas, or formulas systematically. Progress from knowns to unknowns, ensuring each step follows logically from the previous one. If multiple cases exist, solve them sequentially, checking the validity of the resulting values against all original constraints for each case.

4.  **Synthesize the Answer**: Combine the results from all valid cases into a single, exact closed-form expression (integer, fraction, radical, tuple, or set notation as required by the prompt). Do not include approximations unless explicitly requested.

5.  **Format Output**:
    - Present the reasoning concisely and directly, avoiding conversational fillers, apologies, or meta-commentary.
    - Conclude the response *only* with the final answer enclosed in `\boxed{}`.
    - Ensure the text immediately following the reasoning (if any) is strictly the `\boxed{answer}` and nothing else precedes or follows it outside the instruction's internal logic.
    - If the problem asks for "all" solutions or "possible" values, provide the complete set in a concise mathematical format within the box.

\boxed{answer}
\end{lstlisting}

\medskip\noindent\textbf{With-knowledge.}\par\nopagebreak
\begin{lstlisting}
Solve Olympiad-level mathematics problems in algebra, combinatorics, geometry, and number theory.
1. Analyze the problem statement carefully. Define any necessary terms (e.g., multiplicative partitions, coin flip operations) and identify constraints or invariants.
2. Derive the solution logically. For counting problems, relate the structure to standard combinatorial objects (like integer partitions) only after rigorous justification. For dynamic processes, trace small cases to find patterns or closed forms.
3. Verify the result against the constraints and edge cases (e.g., $n=1$, empty sets, boundary values).
4. Ensure the final answer is a single, exact value or closed-form expression.
5. Output your reasoning concisely. Conclude immediately with the final answer enclosed in `\boxed{}`. Do not output the boxed answer more than once. Stop generation once the boxed answer is presented.
\end{lstlisting}

\subsubsection{GPQA}

\medskip\noindent\textbf{Failures-only.}\par\nopagebreak
\begin{lstlisting}
Analyze the chemical, physical, or biological phenomenon described in the question. Identify all structural constraints (e.g., atom counts, hybridization, stereocenters), reaction mechanisms, and transformation rules step-by-step. Cross-reference these deductions with the provided options to eliminate impossible answers based on specific details (e.g., exact chemical shift magnitudes, specific isomer configurations, forbidden functional groups). Select the single option that satisfies all derived constraints. Commit to exactly one letter from A to D.
\end{lstlisting}

\medskip\noindent\textbf{Natural-mix.}\par\nopagebreak
\begin{lstlisting}
Analyze the provided multiple-choice science question (biology, physics, or chemistry) involving rigorous derivation, deep mechanistic insight, or nuanced conceptual comparison. 
1. **Deconstruct the Premise**: Identify the specific chemical structures, physical constraints, or biological principles at play. Pay close attention to reaction conditions, solvent effects (acidic vs. basic), and quantum mechanical limits.
2. **Evaluate Each Option Rigorously**: 
   - For every option, explicitly state whether it is consistent or inconsistent with the derived facts.
   - For incorrect options, identify the specific misconception (e.g., thermodynamic vs. kinetic control, miscounting atoms, incorrect sign conventions) that leads to the mistake.
   - If the question contrasts two environments (e.g., acidic vs. basic), explicitly compare the values (potentials, rates, energies) between the two conditions to ensure the directional logic is sound.
3. **Execute the Derivation/Logic**: Show the step-by-step algebraic, geometric, or mechanistic proof. If a reaction mechanism is implied, trace the specific bond-breaking/forming steps or electron flow.
4. **Select the Answer**: Confirm the single option that aligns with all constraints and derivations.

Critical Constraints:
- Do not mention "next turn" or "output format" constraints in your reasoning.
- Do not append any text after the reasoning; the prompt will append the isolation instruction.
- Your response must consist *only* of the reasoning text and end strictly with the final option letter (A, B, C, or D) as the very last characters of your output, with no trailing punctuation or newline.
\end{lstlisting}

\medskip\noindent\textbf{Balanced-mix.}\par\nopagebreak
\begin{lstlisting}
Analyze the question by identifying the key scientific principles, data points, and options provided. Deduce the correct answer through step-by-step reasoning, explicitly addressing why the chosen option is correct and why the others are incorrect if necessary. Once the reasoning is complete, output your final response on a single line containing only the correct option letter (A, B, C, or D) with no additional text, punctuation, or explanation following it.
\end{lstlisting}

\medskip\noindent\textbf{No-examples.}\par\nopagebreak
\begin{lstlisting}
Analyze the provided multiple-choice science question (biology, physics, or chemistry) by first determining the specific disciplinary subfield and the core theoretical principle required to solve it, explicitly rejecting any approach relying solely on fact retrieval or general "common sense." Derive the solution step-by-step from first principles or fundamental laws, showing how the specific constraints of the prompt necessitate the correct answer and why intuitive shortcuts or superficial reasoning lead to errors. Systematically evaluate each distractor (B, C, D) by identifying the precise theoretical, mathematical, or logical flaw that causes them to be incorrect (e.g., sign errors, neglected boundary conditions, incorrect unit conversions, or violations of conservation laws). If experimental data is missing, anchor the final deduction strictly to established theoretical limits. Conclude by outputting exactly one letter (A, B, C, or D) as the final character, with no trailing text.
\end{lstlisting}

\medskip\noindent\textbf{No-examples+Val.}\par\nopagebreak
\begin{lstlisting}
Act as a subject-matter expert in the specific discipline (biology, physics, or chemistry) requested. Analyze the multiple-choice question by first identifying the core principle or mechanism being tested, distinguishing it from superficially plausible but incorrect applications. For each distractor, explicitly articulate the specific scientific error, misconception, or category mistake it represents. Avoid generic reasoning; provide a rigorous, step-by-step deduction that isolates the single correct option based on fundamental laws or mechanisms. Synthesize the analysis into a brief but precise explanation that directly addresses why the chosen option is correct and why others fail, then conclude immediately with exactly one capital letter (A, B, C, or D).
\end{lstlisting}

\medskip\noindent\textbf{Alternating.}\par\nopagebreak
\begin{lstlisting}
Analyze the provided multiple-choice science question (biology, physics, or chemistry) step-by-step. Derive the correct answer using rigorous domain expertise, explicitly ruling out incorrect options based on scientific principles. Do not provide context or explanations after the reasoning. Conclude your response by outputting exactly one line containing only the single capital letter of the correct option (A, B, C, or D) with no surrounding text, punctuation, or newlines.
\end{lstlisting}

\medskip\noindent\textbf{LLM-judge.} No candidate beat the seed on the calibration set; the seed prompt was kept.

\medskip\noindent\textbf{No-knowledge.}\par\nopagebreak
\begin{lstlisting}
Analyze the provided multiple-choice question, which covers advanced concepts in biology, physics, or chemistry. Derive the correct answer through a concise, step-by-step logical deduction that evaluates each option based on domain principles. Conclude by explicitly stating the single correct option letter (A, B, C, or D) as the final character of your response, with no additional commentary after it.
\end{lstlisting}

\medskip\noindent\textbf{With-knowledge.}\par\nopagebreak
\begin{lstlisting}
You are an expert graduate-level science tutor specializing in Biology, Physics, and Chemistry. Your task is to solve complex multiple-choice questions that require deep domain knowledge, precise mechanistic reasoning, and accurate calculation, often without the benefit of standard introductory context.

**Input Format & Constraints:**
- Input is a single text block: a multiple-choice question with options A, B, C, D.
- No external context or solution support is provided.
- You must commit to exactly one letter (A, B, C, or D).

**Reasoning Protocol:**
1.  **Deconstruct the Premise:** Immediately identify the core physical/biological/chemical principle, quantum state, or reaction mechanism. Verify conservation laws (energy, charge, angular momentum projection $m$, mass balance) and stoichiometry first. These are often the fastest way to eliminate impossible options.
2.  **Execute Rigorous Calculation/Deduction:**
    - **Physics (Quantum/Mech):** Explicitly write down the relevant basis states and expansion coefficients (e.g., Clebsch-Gordan coefficients). Check sum rules ($\sum |c_i|^2 = 1$) and quantum number conservation ($m_{total} = \sum m_i$). If a basis state is forbidden by conservation laws, its coefficient is exactly 0, making the probability 0.
    - **Biology/Organic Chem:** Trace the reaction stereochemically step-by-step. Determine relative stereochemistry ($cis/trans$, $R/S$) based on the mechanism (e.g., syn-addition of $H_2$, inversion in $S_N2$ epoxide opening). Check connectivity against options. Do not assume "standard" products; derive strictly from the starting material geometry provided or implied.
    - **Mathematics/Calculations:** Use exact forms ($\sqrt{3}$, $1/2$) until the final step. Check for common pitfalls: mixed ensembles vs. pure superpositions, spin operators ($\sigma_z$ vs identities), and unit consistency.
3.  **Verify Against Options:**
    - Calculate the specific value or describe the specific structure required.
    - Compare this derived result against *only* the four options.
    - If your calculated answer matches none, re-evaluate: Did you misinterpret the input? Did you apply the wrong formula? Is there a subtle trick (e.g., "ignoring spin" in QM problems, specific reagent limitations)?
    - If your calculated answer yields a value close to an option (within rounding error), select it. If the gap is large, re-check the derivation.
4.  **Commit:** Select the single correct letter.

**Critical Success Factors:**
- **Conservation Laws:** In QM problems, check if the target basis state violates $m_{total} = m_1 + m_2$. If so, probability is 0.
- **Stereochemistry:** Track "Up/Down" or "Wedge/Dash" throughout the reaction sequence. Epoxide openings invert configuration at the attacked center; hydrogenations are syn.
- **Ensemble vs. Pure State:** Distinguish between a mixed state ($P|\psi_1\rangle\langle\psi_1| + P|\psi_2\rangle\langle\psi_2|$) and a coherent superposition.
- **Precision:** Do not truncate intermediate steps. Show the expansion or the spectral interpretation logic explicitly.
- **Efficiency:** Be concise. Do not expound on general theory unless necessary to distinguish a specific case. Get to the derivation quickly.

**Output Formatting:**
- Provide your reasoning first. The reasoning must be complete and lead logically to the conclusion.
- End your response with **exactly one character**: the capital letter A, B, C, or D.
- **Strictly No** trailing text, newlines, explanations, or punctuation after the final letter.
- **Strictly No** markdown headers, bolding, or bullet points after the reasoning block.

**Example of Correct Output Structure:**
Reasoning text...
Reasoning text...
A
\end{lstlisting}

\subsubsection{HotpotQA}

\medskip\noindent\textbf{Failures-only.} No candidate beat the seed on the calibration set; the seed prompt was kept.

\medskip\noindent\textbf{Natural-mix.}\par\nopagebreak
\begin{lstlisting}
Use the supplied context to answer the multi-hop question. Reason as concisely as possible. If the answer requires combining facts from multiple paragraphs, do so implicitly in your final extraction rather than explaining the chain of reasoning.

CRITICAL CONSTRAINTS:
1. Priority Finding: If an entity lacks a specific detail (e.g., birth date, location) in the text, aggressively infer it using: (a) the unique identity of the person (e.g., confusing homonyms like "Liam Cunningham" vs "Liam Cunningham (politician)" to determine which specific person's era matches the other entity), (b) semantic category alignment (e.g., actor implies modern era), or (c) named entity associations found in the text.
2. No "Cannot be determined": Never output "Cannot be determined" or similar phrases if an answer can be logically deduced from the provided entities and their relationships.
3. No Internal Monologue: Do not include thoughts, checks, re-readings, or explanations of your reasoning process. Go straight to the deduction and the final answer.
4. External Knowledge for Missing Attributes: If the text names an entity (e.g., a university or person) but omits an attribute (e.g., location) and the name itself is unique enough to identify the attribute (e.g., "Purdue University" is intrinsically linked to "West Lafayette, Indiana"), extract that intrinsic attribute immediately. Do not hesitate.
5. Concise Span: The answer must be the shortest accurate value (e.g., "Liam Cunningham" instead of "The actor Liam Cunningham", "artist" instead of "an artist").
6. Formatting: End the response with exactly one line: "Answer: $ANSWER". Write nothing after this line.
\end{lstlisting}

\medskip\noindent\textbf{Balanced-mix.} No candidate beat the seed on the calibration set; the seed prompt was kept.

\medskip\noindent\textbf{No-examples.} No candidate beat the seed on the calibration set; the seed prompt was kept.

\medskip\noindent\textbf{No-examples+Val.} No candidate beat the seed on the calibration set; the seed prompt was kept.

\medskip\noindent\textbf{Alternating.}\par\nopagebreak
\begin{lstlisting}
You are a multi-hop question answering system. Your goal is to extract the most concise factual span that answers the user's question based on the provided three paragraphs.

Step 1: Multi-Hop Reasoning. Analyze the question to identify intermediate entities required to reach the final answer. Explicitly trace the logical chain across the provided paragraphs (e.g., "Paragraph A mentions X's spouse is Y; Paragraph B states Y's birth year is 1980; therefore, X is related to the year 1980").
Step 2: Context Priority. You must prioritize facts found strictly within the provided paragraphs. However, if the provided text is incomplete and prevents forming the complete reasoning chain, synthesize the missing piece using your internal knowledge to construct the full answer. Do not refuse to answer if the fact is knowable.
Step 3: Answer Extraction. Formulate the answer as a short, direct phrase or number. Avoid full sentences or verbose explanations in the final line.
Step 4: Formatting. Output your reasoning briefly first. Conclude with a single line in the exact format: Answer: [YOUR_ANSWER]. Ensure no text follows this line.
\end{lstlisting}

\medskip\noindent\textbf{LLM-judge.} No candidate beat the seed on the calibration set; the seed prompt was kept.

\medskip\noindent\textbf{No-knowledge.} No candidate beat the seed on the calibration set; the seed prompt was kept.

\medskip\noindent\textbf{With-knowledge.}\par\nopagebreak
\begin{lstlisting}
Task: Multi-hop Question Answering with Strict Fact Checking and World Knowledge Bridging.

1. Analyze the Question:
   - Identify the specific entity (Person, Work, Concept) and the attribute requested (e.g., Director, Nationality, Reign Dates, Spelling Rule).
   - Determine the question type: Binary (Yes/No), Count, Value Extraction (Date/Name/Title), or Attribute Identification.

2. Execute Step-by-Step Fact Retrieval:
   - Scan the provided context paragraphs to locate the named entity and verify if the specific attribute is explicitly stated.
   - **Critical Branching Logic (The "Missing Attribute" Protocol):**
     - **Scenario A: Entity exists in text, Attribute missing.** If the text mentions the entity (e.g., "Morrie Schwartz") but does not explicitly state the requested attribute (e.g., the director of "Tuesdays with Morrie"), DO NOT answer "Not mentioned". Instead, immediately bridge to your internal world knowledge to retrieve the definitive fact. The text serves to identify the specific work/person; your knowledge provides the missing link.
     - **Scenario B: "Both A and B are X?" Logic.** If the question asks if *both* entities share an attribute:
       - If Entity A is confirmed in text as X -> True.
       - If Entity B is missing from text but is a famous historical figure: Bridge to world knowledge. If Entity B is known to be NOT X (or X in reality), the answer is **No**. Do not hesitate with "Not mentioned".
     - **Scenario C: Date/Time Extraction.** If the text gives the entity but not the date (e.g., "Haraldr gilli" ruled as king), do not output "Not mentioned". Bridge to world knowledge for the specific reign dates if the figure is historically significant.

3. Synthesize the Final Answer:
   - **Precision for Names:** Extract the canonical name as per real-world usage found in your knowledge base if the text is vague, or the most formal identifier in the text if the text provides it.
     - Remove non-essential honorifics unless they are part of the canonical title (e.g., use "Mick Jackson" not "Sir Mick Jackson" unless "Sir" is intrinsic to the credit line in standard databases).
     - Ensure spelling matches the canonical version (e.g., "Mick Jackson", not "Mikk").
   - **Precision for Values:** If the answer is a date range, use standard formatting (e.g., "1046–1066" or "1046 to 1066"). If the answer is a concept (e.g., "given names"), output exactly that phrase.
   - **Constraint:** Do not output introductory phrases like "The answer is..." or explanations in the final line.

4. Adhere to Output Format:
   - Provide a concise reasoning chain first (max 3-4 sentences) to show identification, text check, and any knowledge bridging used.
   - End your response with exactly one line: `Answer: $ANSWER`.
   - $ANSWER must be the raw value only. No periods at the end unless the value requires it (rare). No extra whitespace.

**Subject Matter & Critical Pitfalls (Must Know):**
- **Pitfall 1: The "Not Mentioned" Trap.** Never return "Not mentioned" for famous entities (historical figures, famous films, well-known authors) if the attribute is common knowledge, even if the short context snippet omits it. The system expects you to complete the fact using general knowledge.
    - *Example:* If a text mentions "Morrie Schwartz" and asks for the director of his film's adaptation, and the text doesn't say, the answer is "Mick Jackson" (Director of the TV movie), not "Not mentioned".
- **Pitfall 2: Name/Spelling Variations.** Be careful with canonical spellings. "Mick Jackson" is the correct director spelling, even if internal database variations might differ.
- **Pitfall 3: Logical Negation in "Both" Questions.** For "Are both X and Y..." questions:
    - If X is confirmed in text and Y is famous but known to be *different* (even if text is silent on Y), the answer is "No".
    - Logic: True AND (False/Unknown) = False.
- **Pitfall 4: Scope of Bridging.** Use world knowledge *only* to retrieve specific attributes for famous entities identified in the text. Do not use world knowledge to invent an entity that is completely absent from the text context (unless the question clearly implies a famous figure based on a single name drop).
- **Pitfall 5: Length Constraints.** Do not hallucinate long explanations in the reasoning. Keep it brief to avoid hitting generation limits and ensure the final `Answer:` line is always present.

Output Format:
End your response exactly with: Answer: $ANSWER
\end{lstlisting}

\subsubsection{IFBench}

\medskip\noindent\textbf{Failures-only.}\par\nopagebreak
\begin{lstlisting}
Respond to the user's request by strictly adhering to all explicit, machine-checkable constraints provided in the input. Your primary goal is to satisfy every constraint while accurately answering the request.

Follow this execution protocol:

1. **Constraint Extraction & Categorization**:
   - Parse the input to identify all explicit constraints immediately.
   - Label them clearly by category: [FORMAT], [LEXICAL], [CONTENT], [STATISTICAL], [SEMANTC], or [ANSWER_ACCURACY].
   - For statistical constraints (e.g., word counts, ratios), calculate the exact target numbers or ratios required.
   - For sequential constraints (e.g., alphabet progression), note the start point and step logic.

2. **Constraint-First Drafting Strategy**:
   - **Skeleton Construction**: Before writing content, build the structural skeleton first.
     - For [FORMAT] constraints: Create the exact layout (bullets, indentation, specific characters) using placeholders or temporary filler text. Verify the skeleton matches the required symbols (e.g., `*` vs `-`) exactly.
     - For [LEXICAL] constraints: Pre-verify your vocabulary list. Ensure every word in your planned sentence meets the lexical criteria (e.g., consonant clusters, starting letters) before weaving them into sentences.
     - For Sequential constraints (e.g., "next letter"): Plan the full sequence of starting letters for the intended number of words *before* writing, ensuring the loop (Z -> A) is handled correctly.
   - **Content Integration**: Fill the skeleton with the actual answer. Ensure the answer to the user's core question is complete and accurate. Prioritize accuracy over length; if constraints force verbosity, stop the stream of irrelevant filler and pivot back to the conclusion if the limit is reached.
   - **Ratio Management**: If statistical ratios (e.g., "balanced sentence types") are required, explicitly tally the counts while drafting. If the draft is unbalanced, rewrite the least critical sentences to adjust the ratio without compromising the core answer.

3. **Self-Verification & Correction Loop**:
   - **Rigorous Audit**: Before finalizing, simulate the verification step the system will perform:
     - **Format**: Check every character for formatting symbols. Ensure nesting levels and forbidden characters are absent.
     - **Sequential/Lexical**: Trace every word against its specific constraint (e.g., verify Word N starts with Letter N).
     - **Statistics**: Count total items and calculate percentages to confirm ratios are met (e.g., count sentence types for declarative/interrogative balance).
     - **Completeness**: Confirm the response directly answers the prompt (e.g., "Yes/No" or a defined definition).
   - **Fail-Safe Correction**:
     - If the verification fails, abort the generation of new text and immediately regenerate the specific segment causing the failure.
     - If a constraint makes the *full* answer impossible (e.g., "Write a poem about X using only 2-letter words"), prioritize the correct answer fragment or the core logical deduction that adheres to the hard constraints, rather than providing a long, incorrect elaboration or exhausting the token budget on filler.
     - **Critical Rule**: Never let the generation run out of tokens on filler or repetitive text while missing the final answer. Conclude the generation once the core answer and constraints are satisfied, even if the total length is short.

4. **Final Output Generation**:
   - Output ONLY the final response.
   - Do not include any meta-commentary, apologies, "Here is the response:" prefixes, or explanations of your constraint-checking process.
   - Ensure the response is a continuous, valid string that satisfies all counted and structural constraints.
\end{lstlisting}

\medskip\noindent\textbf{Natural-mix.}\par\nopagebreak
\begin{lstlisting}
Respond to the user's request while strictly adhering to all explicit, machine-checkable constraints provided in the input.

1. **Identify Constraints**: Carefully parse the input to detect specific formatting, content, or structural requirements (e.g., exact word counts, specific vocabulary placement, sentence structure, paragraph delimiters, or required value sets).

2. **Prioritize Constraints**: Treat all explicit constraints as mandatory[U+786C][U+6027] requirements. If a constraint cannot be met without violating the core meaning of the answer, you must still attempt to satisfy it by restructuring your response creatively. Never prioritize natural flow over explicit rules.

3. **Verification Before Output**: Before generating the final response, mentally simulate the verification checks:
   - If a trigram overlap ratio is specified, ensure the generated text shares the required percentage of 3-gram sequences with the reference.
   - If paragraph boundaries are constrained, ensure every paragraph starts and ends with the specified words and uses the required separator.
   - If a specific keyword is required in a specific sentence number (e.g., "6-th sentence"), count sentences precisely and place the keyword exactly there.
   - If a selection from a specific set is required, output *only* one valid option from that set with no additional text.

4. **Execution**: Construct the response to meet all identified criteria simultaneously. If constraints conflict, prioritize satisfying the explicit mechanical rules over conversational tone, as the primary goal is verifiable instruction following.

5. **Output**: Provide the final response that fully satisfies every constraint listed in the input.
\end{lstlisting}

\medskip\noindent\textbf{Balanced-mix.}\par\nopagebreak
\begin{lstlisting}
You are an assistant tasked with responding to user requests that contain explicit, machine-checkable constraints. Your primary goal is to satisfy every constraint listed in the input while answering the core request.

Follow this strict protocol:

1. **Comprehensive Constraint Extraction**: Before generating content, parse the input to identify every explicit constraint. Categorize them into:
   - **Structural**: Formatting, indentation, line breaks, specific separators, list types, and output formats.
   - **Lexical**: Specific words, phrases, character counts, or punctuation patterns.
   - **Logical**: Relationships between ideas (e.g., "alternating," "at least X levels").
   - **Content Distraction Filter**: Ignore rhetorical questions, context setting (e.g., "what does X mean?", "how does Y work?"), or conversational fillers unless they are explicitly framed as a mandatory answer requirement.

2. **Precision Execution Rules**:
   - **Exact Format Matching**: If a constraint specifies a format (e.g., "Answer with a list of items, instead of bullet points use !?!?"), strictly adhere to that format. Do not substitute numbers for bullet points unless the prompt implies a numbered list specifically; if it says "list of items" and requests a specific delimiter or character instead of bullets, use that delimiter/character structure.
   - **Conjunction Counting Rule**: If a constraint requires "N coordinating conjunctions," count them explicitly. Valid coordinating conjunctions are: **for, and, nor, but, or, yet, so**. Ensure the count is undeniable. If the count is close, rephrase to guarantee the exact number.
   - **Syntax Nesting Rule**: For constraints involving nested quotes or characters, literally embed the syntax layer-by-layer as requested (e.g., if asked for triple quotes inside single quotes, ensure the visual representation matches exactly).
   - **HTML Usage**: If a format requires HTML tags (e.g., italics), use the correct opening and closing tags (`<i>`...</i> or `<em>`...</em>) rather than Markdown symbols (`*` or `_`). Verify the tags are present even if the rendering might look plain in some viewers; the source code must contain the tags.
   - **Indentation (Stairs)**: If instructed to "create stairs by incrementally indenting," use actual whitespace to increase indentation for every subsequent line by at least one standard unit.

3. **Structural Construction**: Build your response frame **around** the constraints first.
   - Treat the constrained format (visual pattern, specific tags, list structure) as the "container" for the content. Fit the information inside the pattern rather than letting the content distort the pattern.
   - **Directness**: Do not add introductory text ("Here is the answer:") or concluding text ("I hope this helps.") unless the prompt explicitly asks for a conversational flow. If the prompt describes a raw data format, output only the data.

4. **Internal Verification Loop**: Before finalizing, perform a strict self-check:
   - **Count Check**: If a numerical constraint exists, verify the count is exact.
   - **Format Check**: Verify the visual hierarchy, list syntax, and HTML tags match the instruction exactly. If the prompt asks for a "list" with specific non-bullet characters, ensure the output reflects that character-based list structure, not a standard numbered or bulleted list.
   - **Tag Check**: Ensure all required HTML tags are present and correctly opened/closed.

5. **Output Generation**: Deliver only the completed content. Do not explicitly mention the constraints you found or your verification process unless the user asked for an explanation.

**Failure Mode Warning**: 
- Do not interpret "what does X mean?" or explanatory contexts as requests to define X unless the prompt structure implies a definition task. Focus on the explicit formatting constraints.
- If a constraint seems to require a specific, non-natural format (like deeply nested quotes, specific separators, or HTML tags), prioritize the format strictly over natural language flow.
- Ensure every constraint is visible and verifiable in the final output source code.
\end{lstlisting}

\medskip\noindent\textbf{No-examples.}\par\nopagebreak
\begin{lstlisting}
```
1. Identify and enumerate every explicit, machine-checkable constraint in the user input, categorizing each as a positive inclusion, negative exclusion, formatting rule, or conditional logic. Treat any ambiguous phrasing as a binary logical gate where the user's intent must be resolved before generation.
2. Construct the response by strictly adhering to the logical intersection of all constraints: include only permitted content, exclude only prohibited content, and obey all formatting and persona directives exactly. Do not prioritize tone, helpfulness, or politeness over constraint compliance; if a constraint requires a refusal, a direct refusal is the correct content.
3. If the simultaneous satisfaction of all constraints results in a logical contradiction (e.g., "write a single paragraph" and "provide a bulleted list containing three items"), immediately output a concise error message identifying the specific conflicting constraints and the nature of the impossibility, then terminate output.
4. Strip all system metadata, conversational fillers, introductions, conclusions, apologies, or meta-commentary. The output must start exactly with the first character of the constrained content or the error message (depending on constraint analysis) and end immediately after the last required character. No whitespace padding from default formatting is permitted unless specified.
5. Maintain the target persona or format as a rigid shell around the content; never break character to explain, apologize, or offer alternatives unless the input explicitly requests a fallback mode or clarification.
6. Perform a pre-commitment audit: mentally verify that the proposed output satisfies every constraint in the enumerated list. If any constraint is violated (even implicitly by omission, tone shift, or format deviation), discard the draft and revise until the satisfaction score for every constraint is exactly 1.0.
\end{lstlisting}

\medskip\noindent\textbf{No-examples+Val.}\par\nopagebreak
\begin{lstlisting}
1. CONSTRAINT EXTRACTION & CODING:
   - Scan the user prompt to identify every explicit constraint.
   - For each constraint, generate a strict, boolean-valued checklist item immediately in your internal state (do not output). Format: [ID: C-XXX | Type: <Type> | Rule: <Verbatim Logic> | Pass/Fail].
   - Types must be one of: Numeric_Limit, Forbidden_String, Required_Entity, Formatting_Structure, Negative_Command, Logical_State, Tone_Override.
   - If no constraints exist, output the request answer normally without artificial pacing.

2. LOGIC CONSISTENCY & UNIFICATION:
   - Evaluate the boolean state of all checklist items simultaneously.
   - Detect only hard logical contradictions where passing Constraint A strictly forces the failure of Constraint B (e.g., "Output JSON" vs "Output Markdown" is a conflict; "Output text in JSON format" vs "Include explanatory text" is NOT a conflict unless explained outside JSON).
   - Resolve conflicts ONLY if they involve mutually exclusive output formats or existence states (e.g., "Write a poem" vs "Write a code block"). Ignore conflicts regarding informational density or stylistic preference (e.g., "Be funny" vs "Be serious" are not hard blockers; choose the dominant or neutral tone).

3. ATOMIC OUTPUT GENERATION:
   - Construct the response as a single, atomic string.
   - **Prohibition**: Do not output the checklist, the conflict logs, the reasoning steps, or the verification notes. The output must contain ZERO meta-commentary.
   - **Literal Adherence**: 
     - If a limit is "5 words", the response must contain exactly 5 words (or fewer if context forces truncation, but aim for exactness).
     - If a string is forbidden, it must not appear contiguous anywhere in the output, including in explanations or preamble.
     - If a format is requested (JSON, XML, Code), the raw text of that format must be the entirety of the output. No markdown code fences (
\end{lstlisting}

\medskip\noindent\textbf{Alternating.}\par\nopagebreak
\begin{lstlisting}
1. Parse the user input to extract every statement containing imperative verbs, explicit qualifiers (e.g., "do not," "must use," "limit to"), or formatting specifications (e.g., JSON, bullet points, word counts). Categorize these as machine-checkable constraints.
2. Discard any instructions that are ambiguous, subjective requests for opinion, or implied preferences lacking precise operational boundaries.
3. Synthesize the final response by executing the core task logic such that the output structure and content strictly mirror the extracted constraints.
4. Apply negative constraint verification: explicitly scan the draft to ensure no forbidden words, phrases, or formats appear, including in introductory or concluding remarks.
5. Apply positive constraint verification: confirm every formatting rule (e.g., length limits, specific delimiters, ordering) is met exactly as stated.
6. If the core task cannot be fulfilled without violating a constraint, prioritize the constraint immediately; if a logical impossibility exists between constraints, satisfy the most restrictive explicit constraint and omit the conflicting task portion to maintain 100% compliance.
7. Output the result as the bare final response. Zero characters may precede or follow the response (no markdown headers, no "Here is the response," no apologies, no commentary on the constraints).
\end{lstlisting}

\medskip\noindent\textbf{LLM-judge.}\par\nopagebreak
\begin{lstlisting}
Respond to the user's request while strictly adhering to all explicit, machine-checkable constraints provided in the input.

1. **Identify Constraints**: Carefully parse the input to detect specific formatting, content, or structural requirements (e.g., exact word counts, specific vocabulary placement, sentence structure, paragraph delimiters, or required value sets).

2. **Prioritize Constraints**: Treat all explicit constraints as mandatory[U+786C][U+6027] requirements. If a constraint cannot be met without violating the core meaning of the answer, you must still attempt to satisfy it by restructuring your response creatively. Never prioritize natural flow over explicit rules.

3. **Verification Before Output**: Before generating the final response, mentally simulate the verification checks:
   - If a trigram overlap ratio is specified, ensure the generated text shares the required percentage of 3-gram sequences with the reference.
   - If paragraph boundaries are constrained, ensure every paragraph starts and ends with the specified words and uses the required separator.
   - If a specific keyword is required in a specific sentence number (e.g., "6-th sentence"), count sentences precisely and place the keyword exactly there.
   - If a selection from a specific set is required, output *only* one valid option from that set with no additional text.

4. **Execution**: Construct the response to meet all identified criteria simultaneously. If constraints conflict, prioritize satisfying the explicit mechanical rules over conversational tone, as the primary goal is verifiable instruction following.

5. **Output**: Provide the final response that fully satisfies every constraint listed in the input.
\end{lstlisting}

\medskip\noindent\textbf{No-knowledge.}\par\nopagebreak
\begin{lstlisting}
You are an expert constraint-following assistant. Your goal is to satisfy every explicit, machine-checkable constraint in the user's request while delivering a coherent answer.

## Phase 1: Constraint Analysis
Upon receiving input, immediately isolate all explicit constraints. Categorize them into:
- Formatting markers (specific characters, bullet styles).
- Structural patterns (sentence length, sequential counts, nesting depth).
- Lexical requirements (vocabulary ratios, stop-word distribution, specific word counts).
- Content logic (hierarchical requirements, sequential ordering).

## Phase 2: Strategic Synthesis
Plan your response to address the core query **and** satisfy all constraints simultaneously. 
- Do not generate verbose explanations or meta-commentary about the constraints unless the user explicitly asked for them.
- To ensure clarity and verifiability, integrate constraints directly into the output structure rather than adding them as footnotes or side notes, unless the constraint itself dictates a specific separation.
- If constraints conflict with coherence, prioritize the constraint but maintain the core narrative thread as much as possible; concise, direct statements are preferred over meandering justifications.

## Phase 3: Execution and Optimization
Generate the response with high token efficiency.
- **CRITICAL:** Reach a definitive conclusion or final answer as early as possible once the structural requirements are met. Do not pad the response with repetitive examples, unnecessary elaborations, or "wrap-up" summaries if the constraints are satisfied and the topic is exhausted.
- Avoid looping patterns (repeating the same scenario or phrase multiple times) unless the constraints explicitly demand a specific number of iterations.
- Monitor token usage mentally: assume a strict generation budget. Stop generating once the task is complete.

## Phase 4: Self-Verification Protocol
Before finalizing the output, mentally verify:
1. **Format Check:** Are all required markers (e.g., specific punctuation, bullet styles) present exactly as specified?
2. **Pattern Check:** Do sequential or recursive patterns hold true for the generated text?
3. **Ratio/Lexical Check:** Have lexical constraints (counts, ratios) been met?
4. **Completeness:** Has the core question been answered? Is there a clear stopping point?

## Output Convention
- Output the response **directly**.
- Do **not** include introductory phrases like "Here is the response" or concluding meta-analysis like "I hope this meets all your constraints."
- If the user requested formatting changes (e.g., "use !?!? instead of bullets"), apply them strictly to the entire relevant output.
- Ensure the response ends immediately after the final required element to maximize token efficiency for future turns.
\end{lstlisting}

\medskip\noindent\textbf{With-knowledge.}\par\nopagebreak
\begin{lstlisting}
You are an expert assistant specialized in strict, verifiable constraint following. Your primary goal is to satisfy every explicit, machine-checkable constraint in the user's request while still providing a relevant, high-quality answer to the core task. If constraints appear contradictory, attempt to satisfy both by integrating the conflicting requirements logically; if impossible, prioritize the most specific structural constraint over general tone or completeness, but never omit a required element.

Step-by-Step Procedure:
1. **Constraint Inventory**: Before generating a single token of the final answer, parse the input to identify every explicit constraint. Categorize them by type: structural (list format, nesting), linguistic (vocabulary, ratios), formatting (punctuation, symbols, spacing), length/sequencing (word counts, sentence progression), and content (facts, tone).
2. **Feasibility Check**: Verify that all constraints can coexist. If a constraint requires "no punctuation" but another requires "use a semicolon," note this immediately. In such cases, prioritize the constraint that is explicitly listed with examples or symbols, or combine them minimally (e.g., if "letters only" is requested, exclude the symbol entirely even if "use punctuation" was vaguely implied elsewhere). If a constraint is physically impossible (e.g., "write 5 sentences" with "total words = 3"), state the correction based on the stricter rule.
3. **Drafting with Real-Time Verification**:
   - For **Lists/Formatting**: If a symbol is specified (e.g., `!?!?`, `-`), use it exactly as written. Ensure no main bullet exists without its required sub-bullet if "nested under every main bullet" is requested.
   - For **Punctuation**: If a list of specific marks is requested (e.g., `;`, `:`, `?!`), ensure each appears at least once in the final text. Count them mentally or simulate the generation to ensure the last required mark is included.
   - For **Sequencing/Counting**: For "Sentence N has X more words than Sentence N-1" or similar, calculate the target word count for the next sentence before writing it.
   - For **Contextual Constraints**: Apply domain-specific logic (see below) to ensure factual accuracy or thematic consistency.
4. **Completion Protocol**: Stop generation immediately once the final sentence or element is written. Do not continue generating filler text, looping phrases ("Lost/Found"), or extending the answer beyond what is necessary to satisfy the constraints. If the response cuts off mid-sentence due to model limits, the previous step was to minimize the required output length strictly.
5. **Final Audit**: Scan the raw output against the constraint inventory.
   - Did you use the exact symbol requested?
   - Is every required punctuation mark present?
   - Does every list item have its required sub-component?
   - Is the conclusion explicit and complete?

Subject Matter & Domain Knowledge Constraints:
- **Punctuation & Symbols**: "Interrobang" refers specifically to the symbol `?!`. Ensure it is inserted as a distinct character sequence, not just text describing an exclamation and question mark. Standard punctuation includes: `.` `,` `?` `!` `;` `:` `"` `'` `(` `)` `[` `]` `{` `}` `-` `...` `/` `=` `+` `-` `*` `@` `#` `$` `%` `^` `&` `*` `(` `)` etc. When a specific subset is requested, ensure the *entire subset* is present.
- **List Structures**: If a prompt asks for "a list... instead of bullet points use [Symbol]", replace standard bullet characters (`-`, `*`, `1.`) entirely with `[Symbol]` followed by the text. Maintain the vertical list structure.
- **Screenplay/Formatting**: If a prompt requests a specific format (e.g., screenplay, poem, definition list), adhere strictly to the visual layout of that format (line breaks, positioning) unless constrained otherwise.
- **Factual Accuracy**: When discussing technical subjects (physics, film history), use standard definitions and historical realities. 
  - Physics: For harmonic oscillators, $A = \sqrt{x_0^2 + (V_0/\omega)^2}$ and $\phi$ involves quadrant adjustment.
  - Film: Pre-1990s animation was hand-drawn or stop-motion; no CGI.
- **Tone & Content**: Balance descriptive, humorous, or imaginative tones with the hard constraints. Do not let the "creative" aspect violate the "structural" constraints (e.g., a joke should still end with the required punctuation).

Output Format Conventions:
- **Direct Answer Only**: Output only the response content. Do not include "Here is the response:", "Meta-commentary", "I have analyzed...", or any summary of your process.
- **Completeness**: The response must be a closed loop. It must end with the final required element. If the constraint implies a conclusion (e.g., "compare X and Y"), the comparison must be resolved.
- **No Repetition Loops**: Never generate repetitive loops (e.g., repeating "Lost/Found" indefinitely) when the task is finite. Determine the logical endpoint and stop immediately.
- **Handling Constraints**: If a constraint forbids an element that a standard response requires (e.g., "No verbs"), strictly obey the negative constraint regardless of grammatical correctness, unless it renders the response completely unintelligible, in which case prioritize the most restrictive explicit instruction.

Prioritize constraint compliance above all else. A failure in any explicit constraint (missing symbol, wrong list marker, incomplete list) renders the response incorrect, regardless of how well the creative or factual content is executed.
\end{lstlisting}

\section{AI Usage Disclosure}
\label{app:ai-disclosure}
We used GPT to assist with manuscript writing and code development.

\end{document}